%% file: main.tex
\pdfoutput=1

\documentclass[11pt]{article}

\usepackage[]{acl}

\usepackage{booktabs}       
\usepackage{adjustbox}
\usepackage{multirow}
\usepackage{makecell}
\usepackage{bm}
\usepackage{amsmath}
\usepackage{amsfonts}
\usepackage{bbm}
\usepackage{xspace}
\usepackage{subfigure}
\usepackage{arydshln}

\usepackage{times}
\usepackage{latexsym}
\usepackage{tcolorbox}

\usepackage[T1]{fontenc}

\usepackage[utf8]{inputenc}

\usepackage{microtype}
\usepackage{textcomp}

\definecolor{revision}{rgb}{0, 0.0, 0.0}
\newcommand{\jhk}[1]{\textcolor{black}{#1}}

\def \name{infoVerse\xspace}
\newcommand{\ms}[1]{\tiny{$\pm$#1}}

\title{
\name{}:
A Universal Framework for Dataset Characterization \\ with Multidimensional Meta-information 

}

\author{
        Jaehyung Kim$^\dagger$\thanks{~~This work was done while JK was at Minnesota NLP.} \quad
	Yekyung Kim$^{\ddagger}$ \quad
	Karin de Langis$^{\diamondsuit}$ \quad
	Jinwoo Shin$^{\dagger}$ \quad 
	Dongyeop Kang$^{\diamondsuit}$ \\
	$^\dagger$KAIST, \quad $^\ddagger$Hyundai Motors Company, \quad $^\diamondsuit$University of Minnesota \\
	{\tt jaehyungkim@kaist.ac.kr  dongyeop@umn.edu} 
}
\begin{document}
\maketitle
\begin{abstract}

The success of NLP systems often relies on the availability of large, high-quality datasets.
However, not all samples in these datasets are equally valuable for learning, as some may be redundant or noisy. 
Several methods for characterizing datasets based on model-driven meta-information (\textit{e.g.}, model's confidence) have been developed, but the relationship and complementary effects of these methods have received less attention.
In this paper, we introduce \name{}, a universal framework for dataset characterization, which provides a new feature space that effectively captures multidimensional characteristics of datasets by incorporating various model-driven meta-information. 
\name{} reveals distinctive regions of the dataset that are not apparent in the original semantic space, hence guiding users (or models) in identifying which samples to focus on for exploration, assessment, or annotation.
Additionally, we propose a novel sampling method on \name{} to select a set of data points that maximizes informativeness. 
In three real-world applications (data pruning, active learning, and data annotation), the samples chosen on \name{} space consistently outperform strong baselines in all applications.
Our code and demo are publicly available.\footnote{\url{https://github.com/minnesotanlp/infoVerse}}

\end{abstract}

\input{figures/acl23_figure1}

\section{Introduction}

The construction of large datasets is one of the essential ingredients for success in various NLP tasks \cite{wang2019glue}.
However, not all data points are equally important to learn from; many datasets often contain low-quality samples, \textit{e.g.,} incorrect labels \cite{toneva2018forgetting} or annotation artifacts \cite{gururangan2018annotation}. 
Thus, \textit{data characterization} \cite{roth1990data}, a technique for transforming raw data into useful information for a target task, has a huge potential to improve the model's performance by trimming the problematic samples \cite{pleiss2020aum} or providing better practices for effective data collection, \textit{e.g.}, active learning \cite{beluch2018power} and adversarial annotation \cite{nie2019adv_nli}. 

However, data characterization via human assessment is highly limited due to the huge cost of dealing with a large dataset and the vagueness of the assessment itself.
To this end, several model-driven \textit{meta-information}\footnote{Measurements that quantify implicit information in data points, such as their difficulty to learn.} have been investigated; for example, the model's confidence is a standard meta-information widely used in active learning \cite{beluch2018power}. \citet{swayamdipta2020cartography} recently show that the training dynamics of the model's prediction can indicate the relative importance of training samples.
Various types of meta-information are continuously proposed from different intuitions \cite{salazar2020mlm, paul2021deep}, but their relationship and potential beyond relying on individual one have yet to be explored. 
Hence, this work answers the following two research questions: (1) \textit{is there a (hidden) complementary effect between various meta-information for better data characterization}, and (2) \textit{is the combined meta-information useful for real-world applications?
} 

In this paper, we introduce \textbf{\name{}}: a universal framework for better dataset characterization by incorporating multiple aspects of data characteristics.
To be specific, \name{} combines various types of meta-information which offer the different aspects of data characteristics (e.g., how difficult the sample is to learn, how certain multiple models are, and how likely the sample is). Consequently, we can extract richer information about data informativeness from their complementary effect, and
\name{} could guide users (or models) in what samples to focus on for the exploration, assessment, or annotation.
To extend the advantages of \name{} into real-world problems, we further propose a novel sampling method suitable for \name{} based on determinantal point processes (DPP), which is known to be effective for finding a diverse and high-quality set of samples \cite{gillenwater2012discovering, chen2018fastdpp}. 
It enables us to select data points that maximize the information at a set level rather than a sample level on the multidimensional space of \name{}.

In detail, we first construct \name{} based on the diverse meta-information, which could be broadly classified into four different categories in Section \ref{sec3}.
The complementary effect from the multiple meta-information in \name{} helps reveal distinct regions in the dataset, such as  \textit{hard-to-predict} and \textit{mis-labeled} ones, which are not observable in the original semantic feature space (Section \ref{sec4}).
\jhk{In Section \ref{sec5}, we empirically show that our framework has consistently outperformed the strong baselines in various data-centric applications, like \textit{data pruning} \cite{toneva2018forgetting, paul2021deep}, \textit{active learning} \cite{ Yuan2020ColdstartAL}, and \textit{data annotation} \cite{xie20unsupervised}, although it is not specifically designed for those problems. 
This result opens up the potential of \name{} to other data-centric applications, unlike the application-specified approaches.}

Our results show that a dataset could be distinctively characterized when \textit{many different but complementary dimensions are considered together}.
We believe that our \name{} framework could evolve continuously with the development of new meta-information and hence serve as an effective platform for better characterization of datasets and construction of high-quality datasets.

\section{Related Works}\label{sec2}
\textbf{Quantifying and characterizing dataset.} 
Although the large quantity of datasets is usually got attention for the success of various NLP tasks, the quality of the dataset is also an important factor. 
While constructing data with human-in-the-loop is quite reliable like Dynabench \cite{Kiela2021DynabenchRB}, it is expensive and laboring. 
Hence, some works show the benefits of using a model for quantifying and characterizing datasets; for example, \citet{Rodriguez2021EvaluationEA} demonstrates that the model has the ability to annotate, detect annotation errors, and identify informative examples.
In this line of work, several model-driven \textit{meta-information} have been proposed \cite{toneva2018forgetting, swayamdipta2020cartography, beluch2018power}, as we provide the detailed explanation in Section \ref{sec3} and Appendix \ref{sec:appendix_A}.
Most of the prior works focuses on finding a new meta-information; however, as they are obtained under different intuition and aspects of data characteristics, one can expect the complementary effect between them to provide richer information about the dataset. 
Such direction is under-explored from now on, and we try to fill this gap in this work. 
    
\textbf{Informative subset selection.} 
Finding an informative subset is key for various real-world applications; for example, active learning requires selecting the most informative samples among unlabeled samples for labeling \cite{Settles2009ActiveLL}. 
The most widely used approaches to finding those samples in active learning are based on \textit{uncertainty}.
Although several uncertainty measurements have been successfully applied in various NLP tasks, \citet{Dasgupta2011TwoFO} pointed out that focusing only on the uncertainty leads to a sampling bias with repetitive patterns. 
To this end, \textit{diversity}-based sampling has been explored \cite{sener2018active}.
However, as this approach might select samples that provide little new information, recent works suggest methods combining uncertainty and diversity to take advantage of both methods \cite{Ash2020DeepBA, Yuan2020ColdstartAL}.
Our work provides a better way to select informative samples by effectively incorporating multiple aspects of data characteristics with a single universal framework.

\section{\name{}: Universal Framework for Multi-aspect Data Characterization}\label{sec3}
\input{tables/acl23_table1_features}
\vspace{-1.0mm}

In this section, we present \textit{\name{}}, a universal framework for better data characterization.  
Our high-level idea is extracting the complementary effect between various meta-information, as they are oriented from the different aspects of data characteristics. 
In Section \ref{sec3.1}, we briefly introduce the used meta-information to construct \name{}. 
In Section \ref{sec3.2}, we present a novel sampling method to extend the advantages of \name{} for solving real-world problems. 
We remark that our framework can be easily extended with a new meta-information and not limited to specific ones, while the fixed ones are used for the experiments. 

\subsection{Meta-information for \name{}}\label{sec3.1}
To construct \name{}, one needs to determine \textit{which meta-information to use}, and it is expected to get better capability for data characterization with diverse meta-information.
Hence, we first conduct an extensive investigation of the existing meta-information and find that they could be categorized into four different classes based on how they extract the data characteristics: (1) \textit{Static Measures}, (2) \textit{Training Dynamics}, (3) \textit{Model Uncertainty}, and (4) \textit{Pre-trained Knowledge}. 
In Table \ref{summarization}, we provide the full list of meta-information for each category, and more details are presented in Appendix \ref{sec:appendix_A}. 

\textbf{Static Measures.} 
Meta-information obtained from a single static classifier is arguably the easiest and most natural to use.
\textit{Confidence} to true label and \textit{Entropy} \cite{shannon1948mathematical} of the predictive probability from the classifier's output have been popularly used to characterize each data point.
\textit{BADGE score} \cite{Ash2020DeepBA}, a gradient norm of the linear classifier with respect to training loss, is designed to capture the uncertainty of the prediction. 
Finally, \textit{Task Density} and \textit{Relative Density} defined with kNN distance on the classifier's feature embeddings \cite{devries2020instance} effectively estimates the uniqueness of data at a task level.

\textbf{Training Dynamics.} 
Training dynamics of samples largely varies depending on the samples' characteristics and hence can provide useful information, \textit{e.g.}, the confidence of mislabeled samples slowly increases relative to the normal ones. 
\citet{swayamdipta2020cartography} investigate the usefulness of the mean (\textit{Confidence}) and standard deviation (\textit{Variability}) of the model’s prediction to true class across training epochs; they observe that high variable samples are usually useful and low confident ones have a risk of the noisy label.
\textit{Forgetting Number} \cite{toneva2018forgetting}, a number of the transitions from being classified correctly to incorrectly during training, is also shown to be an effective measurement for finding redundant samples.
\citet{pleiss2020aum} reveals that \textit{Area Under Margin} (AUM), a sum of the gap between the logits of true class and most confusing class across training epochs, is different among easy, hard, and mislabeled samples.

\input{figures/acl23_figure2}

\textbf{Model Uncertainty.} 
As intrinsic randomness within the model affects the samples' prediction, such uncertainty has been widely used in various fields \cite{lakshminarayanan2017simple,lee2021sunrise}.
There are two popular ways to measure the model uncertainty: Monte-Carlo Dropout \textit{(MC-Dropout}) \cite{gal2016dropout} with different Dropout masks for a single model, and \textit{Deep Ensembles} \cite{lakshminarayanan2017simple} with multiple models differently random initialized. 
Specifically, the following four meta-information are used for uncertainty quantification: 1) \textit{Entropy} of the average predicted class distribution of multiple predictions, 2) \textit{BALD} \cite{houlsby2011bayesian}: mutual information between data samples and classifier, 3) \textit{Variation Ratio} \cite{beluch2018power}: a proportion of predictions different with the majority voted one, and 4) \textit{EL2N score} \cite{paul2021deep}: an approximated contribution to the change of training loss. 
In addition, we also include the average and variability of confidence across different models.

\textbf{Pre-trained Knowledge.} 
As general text representation provides complementary information to task's one, using the pre-trained language models to extract meta-information is another popular direction. 
For example, \textit{MLM score} \cite{salazar2020mlm}, a Pseudo-Log-Likelihood \cite{wang2019pll} of Masked Language Model (MLM), gives low values to the sentences with inconsistent context. 
To reduce the computational cost, we use its approximation following \citet{Yuan2020ColdstartAL}. 
Also, \textit{Semantical Density} of each sample based on kNN distance \cite{devries2020instance} using sentence-BERT \cite{reimers2019sentence} can assess its uniqueness compared to other sentences. 

Overall, with 23 meta-information, we construct a new feature space \textit{\name{}}. 
Note that some complementary or redundant meta-information is noticed by low or high correlation values in Figure \ref{fig:fig_confusion}, respectively.
Remarkably, we observe that similar correlations consistently appear across different datasets and models, meaning that this "meta"-information is quite a dataset- and task-agnostic (see Appendix \ref{app:info_other}). 

\subsection{Maximally-Informative Subset Selection}\label{sec3.2}
Consideration of multiple meta-information via \name{} enables better data characterization, but it's non-trivial to apply this framework to practical problems, such as data pruning and active learning. 
One of the main challenges is from the multidimensional nature of \name{}, as it requires a new sampling method rather than existing single-score based sample selections; for example, \citet{beluch2018power, swayamdipta2020cartography} choose top-ranked samples ordered by a specific meta-information like confidence or uncertainty but such ordering is hard to be defined in multidimensional space.
On the other hand, the single-score strategy cannot capture the relationship between the selected samples in the subset; hence, it suffers from the lack of diversity, especially when the size of the subset is small (see Figure \ref{fig:figure_dpp}). 
Lastly, the manual verification of the effectiveness of each feature becomes very costly when multiple features are considered.
Motivated by this, we provide a new effective subset selection method for \name{} based on \textit{determinantal point process} (DPP).
To be specific, we propose to focus on maximizing the informativeness of the subset by leveraging the capability of \name{} for data characterization at a set level; here, DPP provides a way to easily find the effective sampling method by defining the appropriate score and similarity functions.

\input{figures/acl23_figure3}
\input{figures/acl23_figure4}

\textbf{Determinantal point processes.} Formally, a DPP on a set of samples $\mathcal{X}=\{x_1, \dots, x_N\}$ is a probability measure $\mathcal{P}$ on $2^{\mathcal{X}}$, the set of all subsets of $\mathcal{X}$. 
Under the assumption that $\mathcal{P}$ gives a nonzero probability to the empty set, the probability of each subset $X \subseteq \mathcal{X}$ is $\mathcal{P}(X) \propto \text{det}(L_{X})$
where $L \in \mathbb{R}^{N \times N}$ is a real, positive semidefinite (PSD) kernel matrix, and $L_{X}$ denotes the sub-matrix of $L$ which is indexed with the subset $X$. 
Here, we can define the entries of $L$ as follows:
\begin{equation}
    L_{ij} = q(x_i)\phi(x_i)^{T}\phi(x_j)q(x_j)
\end{equation}
where $q(x) \in \mathbb{R}^{+}$ is a \textit{score} of sample $x$ which is used to weight the samples with a high quality or desired property such as high confidence or uncertainty. 
Next, $S_{ij} = \phi(x_i)^{T}\phi(x_j) \in [-1,1]$ represents a \textit{similarity} between samples $x_i$ and $x_j$ with a normalized feature vector $\phi(x) \in \mathbb{R}^{d}, ~||\phi(x)||_{2}=1$. 
We note that the determinant $\text{det}(L_{X})$ is proportional to the volume spanned by the vectors $q(x)\phi(x)$ for $x \in X$, and hence the sets with high-quality and diverse samples have the high probability under distribution from DPP. 
Consequently, DPP provides a natural way to find the maximally-informative subset by selecting the subset with the highest probability among the sets with the same number of samples \cite{kulesza2011kdpp}. 

To apply DPP on \name{}, we consider the following design choices. 
For $S_{ij}$, we use a Gaussian kernel with Euclidean distance \cite{biyik2019batch} on a normalized features $\tilde{x}$ on \name{}:
\begin{equation}
    S_{ij} =  \text{exp}(-\beta ||\tilde{x}_i - \tilde{x}_j||^{2})
\end{equation}
where we use a fixed value $\beta=0.5$.
\jhk{Regarding score $q(x)$, we use a density defined by kNN distance \cite{carbonera2015density} $D_{\text{KNN}}$ to the same class' samples on \name{} for data pruning: 
\begin{equation}
    D_{\text{KNN}}(x) = - \min_{K}\{||\hat{x} - x||_{2} ~|~ \hat{x} \in \mathcal{X} \}
\end{equation}
where $\min_{K}{\{\cdot}\}$ is defined as the $K$th smallest value in a set. 
In our experiments, we commonly set $K = 5$.
As $D_{\text{KNN}}$ has a negative value, we use its negative inverse to define a positive score function, \textit{i.e.}, $q(x) = -1/D_{\text{KNN}}$.}
Intuitively, it encourages selecting the samples that preserve the informativeness captured on \name{} as much as possible. For active learning {\color{revision} and data annotation}, we use its inverse as $q(x)$ to select samples that have information hard to be captured by their neighbor and hence will be beneficial when labeled.
Finally, we adopt the efficient greedy method \cite{chen2018fastdpp} for DPP, as finding a set with the highest probability is NP-hard. 

Figure \ref{fig:figure_dpp} shows the {\color{revision} 10} selected samples with different selection methods on the QNLI training dataset with a fine-tuned RoBERTa-large classifier. 
Here, one can observe that single-score based selection methods like \textit{Ambig} and \textit{Hard} \cite{swayamdipta2020cartography} actually suffer from the lack of diversity. 
CoreSet \cite{sener2018active} or K-means clustering can select diverse samples, but they are known to be vulnerable to the existence of outliers \cite{georgogiannis2016robust}.
In contrast, DPP successfully selects informative and diverse samples; as shown in the right below of Figure \ref{fig:figure_dpp}, the log determinant with DPP, i.e., approximate set-informativeness, is much higher than the others.   

\section{Dataset Characterization via \name{}}\label{sec4}

In this section, we demonstrate \textit{how \name{} could help analyze a given dataset via better data characterization.} 
Specifically, Figure \ref{fig:figure2} presents \name{} on QNLI dataset\footnote{Natural language inference dataset derived from SQuAD.} along with other representative feature spaces for data characterization: \textit{classifier's embedding} (at final layer before linear head) and \textit{data map} \cite{swayamdipta2020cartography}. 
As the classifier's embedding and \name{} are high dimensional space, we project them to 2D space via t-SNE \cite{van2008tsne} for visualization.
First, one can observe that \name{} maps the samples into distinguishable regions based on their characters; for example, samples with high \textit{variability} are further mapped to some different regions.
To be specific, in Figure \ref{fig:figure2}, they have high variability in a common while having a difference in other measures: the regions with dotted boxes have relatively high \textit{Ensemble Entropy} and \textit{MC Entropy}, respectively, which cannot be distinguishable in the data map, showing the advantage of \name{} in dataset characterization.

\input{tables/acl23_table2_pruning}

This benefit of \name{} for dataset characterization is more clear when we focus on the incorrectly predicted samples (black squares/circles in Figure \ref{fig:figure2}).
As shown in the left of Figure \ref{fig:figure2}, it is hard to find their characteristics on the classifier's embedding as they are scattered over different regions.
Data map \cite{swayamdipta2020cartography} maps these samples to regions with low confidence (hard-to-learn) or high variability (ambiguous), but not perfectly as these regions also include correctly predicted samples.
In contrast, \name{} successfully characterizes these incorrectly predicted samples and maps them into three distinct regions with a different distribution of meta-information: 1) \textit{Hard-and-disagreed}, 2) \textit{Easy-to-mistake}, and 3) \textit{Ambiguous}. 
As shown in the right of Figure \ref{fig:figure2}, both \textit{Hard-and-disagreed} and \textit{Ambiguous} regions have high ensemble uncertainty, but \textit{Hard-and-disagreed} region has relatively low confidence and variability which means that it is also \textit{hard to learn}.
It might imply its incorrect prediction due to intrinsic difficulty as one can verify in the examples in Figure \ref{fig:figure2}.
In contrast, \textit{Easy-to-mistake} region has much lower uncertainty than other incorrectly predicted regions, which indicates the prediction is certainly wrong.
It might indicate that the mistakes happened during annotation even though they are easy ones to annotate correctly.
More results of data characterization on other datasets with \name{} are presented in Appendix \ref{app:info_other}. 

\section{infoVerse for real-world applications}\label{sec5}
In this section, we demonstrate the advantages of \name{} on three real-world problems: 1) \textit{Data Pruning} \cite{swayamdipta2020cartography}, 2) \textit{Active Learning} \cite{beluch2018power}, and 3) \textit{Data Annotation} \cite{kiela2021dynabench}.
These problems commonly \textit{require characterizing and quantifying data to determine which samples to select}, but with different goals; hence, various methods have been explored separately. 
In contrast, we will demonstrate the potential of \name{} as a universal framework to deal with such data-centric problems. 

\subsection{Data Pruning}\label{sec5.1}

The goal of data pruning is selecting the most informative subset of a given training dataset while keeping the performance of the model trained on the subset; hence, measuring the sample's informativeness becomes a key for data pruning. 
This problem has been popularly investigated with various meta-information as an important problem for improving the efficiency of various NLP tasks.

\textbf{Setups.} For the experiments of data pruning, we first use two datasets, QNLI \cite{wang2019glue} and WinoGrande \cite{Sakaguchi2020WINOGRANDEAA}, following the recent work \cite{swayamdipta2020cartography}.
Then, we use three additional datasets,  SST-2 \cite{Socher2013RecursiveDM}, CoLA \cite{Warstadt2019NeuralNA}, and RTE \cite{wang2019glue}, for the comprehensive demonstration.
We consider 8 different pruning ratios (\{17\%, 34\%, 50\%, 67\%, 75\%, 83\%, 87\%, 91\%\}), which includes more challenging setups compared to the previous works \cite{swayamdipta2020cartography, paul2021deep}.
We run all experiments by fine-tuning RoBERTa-large \cite{liu2019roberta}, following \cite{swayamdipta2020cartography}. 

To demonstrate the effectiveness of \name{}-DPP, we compare it with various state-of-the-art approaches to data pruning. 
We first consider a random-sampling (\textit{Random}); then, we consider three different approaches in \cite{swayamdipta2020cartography} (\textit{Easy}, \textit{Hard}, and \textit{Ambig}), which selects the samples by scoring them with a specific meta-information (average confidence and variability).
In addition, we introduce two additional data pruning works: \textit{Forget} \cite{toneva2018forgetting} and \textit{EL2N} \cite{paul2021deep}.
Finally, we consider density-based approaches as they are arguably the most natural ways to preserve the characteristics of the dataset: \textit{Coreset} \cite{sener2018active} and \textit{Density} \cite{Yuan2020ColdstartAL}. 
More details of datasets and training can be found in Appendix \ref{app_pruning}.

\input{figures/acl23_figure5}

\textbf{Results.} 
Figure \ref{fig:fig_main_pruning} shows the performance under varied pruning ratios on WinoGrande (see Appendix \ref{app_more_results} for the other tasks). 
We first note that the effectiveness of each pruning method significantly varies on the pruning ratio.
For example, \textit{Hard} and \textit{Ambig} show good performance at small pruning ratios, but they often fail at large pruning ratios, similarly observed in \cite{swayamdipta2020cartography}.
On the other hand, density-based methods are robust to the varied pruning ratios, although overall performance is relatively low. 
Hence, to compare each baseline by considering all the pruning ratios together, we compare a single averaged performance in Table \ref{table:pruning} similar to Area Under Curve \cite{tai1994mathematical}. 
Here, one can observe that \name{} with DPP (\name{}-DPP) consistently outperforms other pruning methods across all datasets. 

\input{tables/acl23_table3_pruning_ablation}
\input{figures/acl23_figure6}
\input{tables/acl23_table4_active}

To demonstrate the advantages of \name{}-DPP, we present Figure \ref{fig:fig_confusion_pruning} to qualitatively show how the proposed method works for data pruning.
Interestingly, types of majority meta-information of selected samples dynamically change as the pruning ratio increases, from \textit{confidence} to \textit{model uncertainty} to \textit{variability}. 
After the most redundant samples (\textit{i.e.}, high-confidence) are pruned, followed by hard and uncertain samples. 
In contrast, other baselines do not have any pattern for selection or just have static ones, as shown in Figure \ref{fig:app_pruning_analysis2}.

It is worth noting that effective pruning strategies would indeed vary with the pruning ratio or given; for instance, \cite{swayamdipta2020cartography} disclose that high-confidence samples should be pruned at low pruning ratios due to redundancy, but these samples become essential for training as the ratio increases (e.g., 83\%). 
While \cite{swayamdipta2020cartography} could manually check the varied effectiveness of confidence and find the effective pruning strategy based on that, such a manual approach becomes very costly when the number of considered measurements increases. 
In this aspect, our framework offers an efficient solution as it prunes the samples toward maximizing the informativeness of the remaining samples, rather than focusing on specific measurements. 
Also, the observed pruning curriculum demonstrates how \name{} with DPP actually outperforms the other selection methods, by automatically adapting the pruning strategy across varying ratios. 

In addition, to verify the complementary effect between different categories of meta-information, we compare the model's performance by pruning the samples based on each category using DPP. 
As shown in Table \ref{table:ablation_info_feature}, the effectiveness is largely different depending on whether each category can capture the important aspect of the dataset for data pruning. 
However, when they are combined to construct \name{}, the performance is significantly improved which implicitly reveals that they are mutually complementary. 
More results are presented in Appendix \ref{app_more_results} and \ref{app_table2}.

\subsection{Active Learning}\label{sec5.2}

Active learning (AL) is a task that finds the most informative subset from unlabeled samples when labeled and used for training the model.
AL usually consists of multiple iterations of the following two steps: (1) \textit{select} a subset of unlabeled data under a specific sampling method and expand the labeled data by annotating the subset. 
(2) Then, \textit{train} the model with the new training dataset.  
More details and experimental results are in Appendix \ref{app_al_method}.

\textbf{Setups.} To demonstrate the effectiveness of \name{} in AL, we compare it with the state-of-the-art AL methods on the various datasets, following the recent works of AL for NLP tasks \cite{Yuan2020ColdstartAL, margatina2021active}.
Specifically, we evaluate \name{} on three datasets: SST-2, RTE, and AGNEWS \cite{Zhang2015CharacterlevelCN}.
Also, several AL methods are used as the baselines, which are based on three different strategies: (1) uncertainty-based (\textit{Entropy} and \textit{BALD}, as described in \S\ref{sec3.1}), (2) diversity-based (\textit{BERT-KM} and \textit{FT-BERT-KM} \cite{Yuan2020ColdstartAL}) which focus to cover data distribution, and (3) hybrid method to consider both aspects jointly (\textit{BADGE} \cite{Ash2020DeepBA} and \textit{ALPS} \cite{Yuan2020ColdstartAL}), and \textit{Random} sampling.
All experiments are conducted using BERT-base \cite{devlin2019bert}. 
We construct \name{} of unlabeled datasets with their pseudo-labels \cite{lee2013pseudo}.

\input{figures/acl23_figure7}
\input{figures/acl23_figure8}

\textbf{Results.} 
We first summarize the results in Figure \ref{fig:fig_app} and Table \ref{table:al}.
Figure \ref{fig:fig_app} presents the test accuracy of the trained model at each AL iteration on SST-2 dataset (the results of other datasets are presented in Appendix \ref{app_more_results}, due to limited space). 
In addition, Table \ref{table:al} shows the average test accuracy across multiple AL iterations and implies the overall performance of each method. 
Here, one can observe that \name{}-DPP shows consistent improvements over other baselines; \name{}-DPP outperforms the baselines in RTE and SST-2, while it shows comparable performance with the highest performing baseline \textit{BALD} in AGNEWS. 
Consequently, \name{}-DPP achieves the lowest average rank ($1.3$) among the tested AL methods. 

Next, we conduct additional experiments to understand in depth how \name{}-DPP selects the informative unlabeled samples and improves the model's performance.
Specifically, on SST-2, we compare the average of meta-information of the selected samples by \name{}-DPP and two representative baselines, \textit{Random} and \textit{Entropy}.\footnote{\textit{Random} is the most simple approach and \textit{Entropy} shows the second best performance.}
Figure \ref{fig:active_analysis} presents the results;  \textit{Entropy} selects the mostly uncertain samples (\ref{fig:1c}), but it relatively suffers to select the unseen samples (\ref{fig:1d}) and also has a risk to select noisy samples (\ref{fig:1a}). 
In contrast, \name{}-DPP incorporates the multiple meta-information during the selection; for example, it selects the mostly variable samples with moderately low confidence, which has been demonstrated as a key characteristic for effective training samples \cite{swayamdipta2020cartography}. 
Also, the selected samples capture a certain level of uncertainty along with a low sentence-level density (\textit{i.e.}, hence can introduce the new pattern in training samples).

\subsection{Data Annotation}\label{sec5.3}

\input{figures/acl23_figure9}
\input{tables/acl23_table5_real_annotation}

Finally, we demonstrate the advantage of \name{} on data annotation \cite{kiela2021dynabench}, to provide the most effective set of unlabeled samples that are expected to improve the model's performance after they are annotated with \textit{human labelers}.

\textbf{Setups.}
We consider two datasets, SST-5 \cite{Socher2013RecursiveDM} and IMP datasets \cite{du2021self}.
Following \cite{du2021self}, we first conduct an unsupervised data retrieval to prepare high-quality 10,000 candidates among 2M unlabeled sentences from Common Crawl \cite{wenzek2020ccnet} and Reddit corpus.\footnote{\url{https://convokit.cornell.edu/documentation/subreddit.html}}
We then apply each selection method to choose final queries for data annotation: 1,000 samples for SST-5 and 600 samples for IMP, respectively. 
Finally, we ask crowd-workers to annotate the selected samples using Amazon's Mechanical Turk \cite{crowston2012amazon} with at least three different annotators.
We compare the two representative methods, \textit{Random} and \textit{Entropy}, with ours (\textit{\name{}-DPP}) due to the limited resources.
We include more details in Appendix \ref{app_annotation}.

\textbf{Results.} 
Table \ref{table:data_annotation} shows the performance with different selection methods on SST-5 and IMP datasets. 
One can observe that \name{} with DPP consistently finds more informative sets of samples leading to extra performance gain than the other sampling methods on both datasets.
We further measure disagreement between annotators on the newly-annotated dataset in the IMP task in Figure \ref{fig:figure_disagree}.
The order of annotated samples by ours is more linearly aligned with the annotators' disagreement than other sampling methods, indicating that our method prefers to choose more hard and informative samples first.
Consequently, unlike the prior methods relying on single meta-information like confidence \cite{xie2020self} or uncertainty \cite{mukherjee2020uncertainty}, our multi-dimensional approach with \name{} could provide useful contributions for data annotation.
\jhk{Finally, we remark that experimental results and relevant discussions about computational cost and complementary effect of meta-information are presented in Appendix \ref{app_table2} and \ref{supp.cost}, respectively.}

\section{Conclusion}

We propose a new framework, \name{} to characterize the dataset in various aspects of data informativeness.
\jhk{To be specific, \name{} utilizes various types of meta-information which offer different aspects of data characteristics.} 
The combination of diverse meta-information helps detect distinct regions of dataset characteristics, which are not observable in the previous feature spaces.
\jhk{In addition, we further propose a novel sampling method to select data points that maximize the information at a set level rather than a sample level on the multidimensional space of \name{}.}
We empirically demonstrate the benefit of \name{} on three applications: data pruning, active learning, and data annotation.
\name{} with the proposed subset selection method shows consistent improvement over the strong baselines of each problem.
We believe our framework will emerge with the growth of data-centric approaches and contribute to a better understanding of the dataset and improvement of the dataset's quality.

\section*{Limitations}

In this paper, we propose a new framework that extracts the various aspect of information about given data, relying on the existing model-driven meta-information from the trained models.
Hence, if there are some flaws within the used models, such as biased prediction \cite{sun2019mitigating} or learning of spurious correlation \cite{liu2021just}, then our framework can be directly affected and may have a risk of inheritance or amplification of such problematic behaviors. 
However, as our framework is not limited to any specific models and meta-information, one can prevent this problem by using the robustly trained models \cite{sagawa2020distributionally} or introducing more specialized meta-information \cite{lee2021self} for these problems.  
In addition, despite the empirical gains we find, our subset selection method is not theoretically guaranteed to be (or tightly bound to) the optimal set of max informativeness, which remains an interesting direction. 
A further study is necessary showing that selected samples from \name{} could lead to low inter-annotator agreement in manual annotation but provide more accurate information than pseudo-labels.
Abnormality detection using \name{}, like noisy labels, out-of-distribution, or annotation artifacts, could be interesting future directions.

\section*{Broader Impact and Ethical
Implications}
Our work aims to quantify the data informativeness with multi-perspective for capturing properties that can not be revealed by a single perspective. Especially, \name{} lends some insight into data by models what we have. Thus, \name{} has the potential for guiding the construction of high-quality datasets, \textit{e.g.,} removing mis-labeled samples. 
From these points, it is possible to develop a system or general platform for effectively collecting data like Dynabench\footnote{\url{https://dynabench.org/}} and Snorkle\footnote{\url{https://www.snorkel.org/}}. We anticipate that the general platform of \name{} could be contributing to human-involved machine learning systems.

Although our work empirically demonstrates the improvement over various real-world problems, the current version of \name{} has a potential risk to be vulnerable to sample undesirable properties (e.g., gender bias \cite{Bordia2019IdentifyingAR}) in a dataset, as we construct \name{} with meta-information measure do not consider such properties. However, it can be easily alleviated by adding various measurements which represent 'fairness' thanks to the extensibility of our framework. Hence, we believe that our proposed method can be personalized to the purpose of data collection.   

\bibliography{acl,anthology}
\bibliographystyle{acl_natbib}

\clearpage
\appendix

\section{Summary and Formal Definition of Meta-information}
\label{sec:appendix_A}
\input{tables/acl23_table6_features_details}

In this section, we first present a detailed summarization of considered meta-information in Table \ref{citation-guide}. 
Then, we provide a formal definition of each \textit{meta-information} introduced in Section \ref{sec3.1}.
Here, we consider a classification task with $K$ classes for the explanation. 
$x$ and $y$ indicate the input token and the corresponding true label, respectively. 
$f_{\boldsymbol{\theta}}$ indicates the classifier, which is pre-trained Transformer \cite{vaswani2017attention} such as BERT \cite{devlin2019bert} or RoBERTa \cite{liu2019roberta}.
$p_{\boldsymbol{\theta}}=\text{Softmax}(f_{\boldsymbol{\theta}})$ is a predictive distribution of classifier and $z_{\boldsymbol{\theta}}$ is a contextualized embedding before linear classifier in $f_{\boldsymbol{\theta}}=W^{T} z_{\boldsymbol{\theta}}$. 

\subsection{Static Measures} 
Static measures are the meta-information extracted from a single static model, which is the most natural and easy way. In total, \textbf{5} different meta-information is used.
\\
\textbf{1. Task Density} \cite{devries2020instance}

Here, Euclidean distance to $K$th nearest sample is used as density following \citet{carbonera2015density}. 
\begin{equation*}
    D_{\text{KNN}}(x) = - \min_{K}\{||\hat{z}_{\boldsymbol{\theta}} - z_{\boldsymbol{\theta}}(x)||_{2}\}
\end{equation*}
where $\hat{z}_{\boldsymbol{\theta}} \in D_{\tt train} \backslash \{ z_{\boldsymbol{\theta}}(x) \}$ and $\min_{K}{\{\cdot}\}$ is defined as the $K$th smallest value in a set. In our experiments, we set $K$ = 5. 
\\
\textbf{2. Relative Density} \cite{devries2020instance}

As the \textbf{Task Density} does not utilize the label information, we further consider the relative density which is the difference of kNN density to true class samples and other class samples. 
Hence, if this value is large, it implies that $x$ is near to the true class and far from other classes. 
In our experiments, we set $K$ = 5.
\\
\textbf{3. Static Confidence}
\begin{equation*}
    \bar{\mu}(x) = p_{\boldsymbol{\theta}}(y|x)
\end{equation*}
\\
\textbf{4. Static Entropy} \cite{shannon1948mathematical}
\begin{equation*}
    \bar{H}_{\tt Ent}(x) = - \sum_{k=1}^{K} p_{\boldsymbol{\theta}}(k|x) \cdot \log p_{\boldsymbol{\theta}}(k|x)
\end{equation*}
\\
\textbf{5. BADGE} \cite{Ash2020DeepBA}
BADGE is originally proposed for active learning, to select the diverse and uncertain samples.
\begin{equation*}
    s_{\tt BADGE}(x) = ||(p_{\boldsymbol{\theta}}(x) - y) \cdot z_{\boldsymbol{\theta}}(x)||_{2}
\end{equation*}
\\

\subsection{Training Dynamics} 
Training dynamics of samples largely varies depending on the samples’ characteristic and hence can provide useful information, e.g., the confidence of mislabeled samples slowly increases relative to the normal ones. We totally find \textbf{4} corresponding meta-information in this category. Here, $E$ is the total training epoch.
\\
\textbf{6. Average Confidence} \cite{swayamdipta2020cartography}
\begin{equation*}
    \hat{\mu}(x) = \frac{1}{E} \sum_{e=1}^{E} p_{\boldsymbol{\theta}^{(e)}}(y|x)
\end{equation*}
\\
\textbf{7. Variability} \cite{swayamdipta2020cartography}
\begin{equation*}
    \hat{\sigma}(x) = \sqrt{\frac{\sum_{e=1}^{E} \big(p_{\boldsymbol{\theta}^{(e)}}(y|x) - \hat{\mu}(x)\big)^{2}}{E}}
\end{equation*}
\\
\textbf{8. Forgetting Number} \cite{toneva2018forgetting}
\begin{equation*}
    n_{\tt forget}(x) = \sum_{e=1}^{E} \mathbbm{1}(\text{acc}^{(e)}_{i} > \text{acc}^{(e)}_{i+1})
\end{equation*}
where $\text{acc}^{t}(x) = \mathbbm{1}\big(\arg \max_{k} p_{\boldsymbol{\theta}^{(e)}}(k|x) = y \big)$
\\
\textbf{9. Area Under Margin} \cite{pleiss2020aum}
\begin{equation*}
    \text{AUM}(x) =\frac{1}{E} \sum_{e=1}^{E} M^{(e)}(x,y)
\end{equation*}
where $M^{(e)}(x,y) = f_{\boldsymbol{\theta}^{(e)}}(y|x) - \max_{k \ne y} f_{\boldsymbol{\theta}^{(e)}}(k|x)$
\\
\subsection{Model Uncertainty} 
As intrinsic randomness within the model affects the samples’ prediction, such uncertainty has been widely used in various fields. Total \textbf{6} different meta-information are considered. As we consider the ensemble from MC-Dropout and the ensemble of multiple random seed models, total \textbf{12} measures are considered.
Here, $T$ is the total number of models trained with different random seeds.  
\\
\textbf{10. EL2N score} \cite{paul2021deep}
\begin{equation*}
    s_{\tt EL2N}(x) = \sum_{t=1}^{T} || p_{\boldsymbol{\theta}^{(t)}}(x) - y||_{2}
\end{equation*}
\\
\textbf{11. Ensemble Entropy} \cite{shannon1948mathematical}
\begin{equation*}
    H_{\tt Ent}(x)= - \sum_{k=1}^{K} p_{\tt avg}(k|x) \cdot \log p_{\tt avg}(k|x)
\end{equation*}
where $p_{\tt avg}(k|x) = \frac{1}{T} \sum_{t=1}^{T} p_{\boldsymbol{\theta}^{(t)}}(k|x)$
\\
\textbf{12. BALD} \cite{houlsby2011bayesian}
\begin{equation*}
\begin{aligned}
    & I_{\tt BALD}(x) = H_{\tt Ent}(x) \\ 
    & - \frac{1}{T} \sum_{t=1}^{T} \sum_{k=1}^{K} - p_{\boldsymbol{\theta}^{(t)}}(k|x) \cdot \log p_{\boldsymbol{\theta}^{(t)}}(k|x)
\end{aligned}
\end{equation*}
\\
\textbf{13. Variance ratio} \cite{beluch2018power}
\begin{equation*}
    v(x)= 1 - \frac{f_m(x)}{T}
\end{equation*}
where $f_m(x) = \sum_{t=1}^{T} \mathbbm{1}(\arg \max_{k} p_{\boldsymbol{\theta}^{(t)}}(k|x) = \hat{y}_{\tt avg}(x), ~ \hat{y}_{\tt avg}(x) = \arg \max_{k} p_{\tt avg}(k|x)$
\\
Furthermore, we consider \textbf{14. Ensemble Confidence} and \textbf{15. Ensemble Variability} which just changes the role of epoch $E$ to the number of models $T$. Also, we further consider the same meta-information by simulating the ensemble with Monte-Carlo dropout (MC-dropout) \cite{gal2016dropout}, \textit{i.e.}, using different Dropout masked models for the ensemble. From this, we obtain \textbf{16. MC EL2N score} to \textbf{21. MC Ensemble Variability}.  
\\

\subsection{Pre-trained Knowledge}
As general text representation provides complementary information to task’s one, using the pre-trained language models to extract meta-information is another popular direction. 
We use \textbf{2} meta-information measures extracted from pre-trained models which are agnostic to the target task.
\\
\textbf{22. Semantical Density}
\cite{devries2020instance}
\begin{equation*}
    D_{\text{KNN}}(x) = - \min_{K}\{||\hat{z}_{\boldsymbol{\theta}} - z_{\boldsymbol{\theta}}(x)||_{2}\}
\end{equation*}
where $\hat{z}_{\boldsymbol{\theta}} \in D_{\tt train} \backslash \{ z_{\boldsymbol{\theta}}(x) \}$ and $\min_{K}{\{\cdot}\}$ is defined as the $K$th smallest value in a set. We set $K$ = 5. Unlike \textbf{ Task Density}, $z$ is extracted from a pre-trained sentence encoder, e.g., sBERT \cite{reimers2019sentence} which is known to capture the relationship between sentences.
\\
\textbf{23. Pseudo-Log-Likelihood (PLL)} \cite{salazar2020mlm, Yuan2020ColdstartAL}

Originally, \citet{salazar2020mlm} use the following masked language modeling (MLM) score from pre-trained language model $\boldsymbol{\theta}^{\tt MLM}$ as Pseudo-Log-Likelihood (PLL) \cite{wang2019pll}.
\begin{equation*}
    \text{PLL}(x) = \sum_{l=1}^{L} \log p_{\boldsymbol{\theta}^{\tt MLM}}(x_{l}|x_{ \backslash l}) 
\end{equation*}
where $x_{\backslash l} := (x_{1},\dots,x_{l-1}, x_{l+1}, x_{L)}$. However, it requires $L$ times of inference for calculation. Hence, we instead its approximation following \citet{Yuan2020ColdstartAL}, which just calculates PLL at once without masking tokens. \\

\section{Details of Experiment}\label{AppendixB}

\subsection{Dataset}
\input{tables/acl23_table7_datasets}
In this section, we provide details of all datasets we used in this work and hyperparameters used for training the models. 
For all datasets, we used the given standard training and validation sets. 
We present the details of downstream datasets in Table \ref{tab:dataset}. 
All of the data we used can be downloaded from HuggingFace dataset \url{https://huggingface.co/datasets/}. 
For experiments, we report accuracy using the official test set on WinoGrande and AGNEWS.
For those where the label for the official test set is not available (SST-2, RTE, MNLI, and QNLI), we use the given validation set. 
Also, the maximum sequence length is commonly set to 128.

\subsection{Data Pruning}\label{app_pruning}

For data pruning experiments, we commonly fine-tune RoBERTa-large classifier \cite{liu2019roberta} which has the 355M parameters, following \cite{swayamdipta2020cartography}. 
For fine-tuning, we commonly train it for 10 epochs with learning rate 1e-5 and batch-size 16 (except WinoGrande with 64 following \cite{swayamdipta2020cartography} due to the optimization difficulty) with Adam optimizer \cite{kingma2014adam}. 
For each pruning ratio, selection method, and dataset, we run three times with different random seeds.

\subsection{Active Learning}
\label{app_al_method}

Active learning (AL) is a task that finds the most informative subset from unlabeled samples when they are labeled and added to the training dataset.
AL usually consists of multiple iterations of the following two steps: (1) annotates a subset of unlabeled data chosen by a sampling method, and (2) adds the labeled data to the previous round of the dataset and re-train the model with the new training dataset.
For each round, we trained the model from scratch to avoid overfitting, following \citet{Hu2019ActiveLW}.

To be specific, for the experiments in Section \ref{sec5}, we select 100 examples for RTE and 500 examples for CoLA and AGNEWS for each iteration from the training dataset, respectively.\footnote{The initial (\textit{i.e.}, the first iteration) labeled samples of all AL methods are commonly selected by random sampling.} Note that the specific selection 
Then, they are moved to the labeled dataset from the unlabeled pool in each iteration.  
 
To simulate AL, we sample a batch of $k$ sentences from the training dataset, query labels for this batch, and  
Batch size $k$ is set to 500 for SST-2 and AGNEWS, and 100 for RTE which is a relatively small dataset.
For each sampling method and dataset, we run an AL simulation five times with different random seeds.
Also, we fine-tune models on five epochs for SST-2 and AGNEWS, and ten epochs for the RTE dataset.  
We experiment with the BERT-base model which has 110M parameters provided by HuggingFace Transformers \cite{Wolf2019HuggingFacesTS} with Apache License 2.0.
Our implementation is based on existing code repositories\footnote{\url{https://github.com/forest-snow/alps}} with MIT License and used the same hyper-parameter\cite{Yuan2020ColdstartAL}. 
We use AdamW \cite{Loshchilov2019DecoupledWD} with a learning rate of 2e-5. 
\\
\textbf{BERT-KM} \cite{Yuan2020ColdstartAL}: As a diversity-based baseline, applying k-means clustering to the l2 normalized BERT output embeddings of the fine-tuned model to select $k$ data points. 
\\
\textbf{FT-BERT-KM} \cite{Yuan2020ColdstartAL}: Using the same algorithm as BERT-KM except for the BERT embeddings from the previously fine-tuned model are used.
\\
\textbf{ALPS} \cite{Yuan2020ColdstartAL}: Input sentence is randomly masked, then predict the masked language model(MLM) loss of BERT as a proxy for model uncertainty

\subsection{Data Annotation}
\label{app_annotation}
\input{figures/acl23_figure10}
\input{figures/acl23_figure11}

Here, we provide the details about the annotation pipeline with crowd workers. 
During experiments, we annotate the selected unlabeled samples with each selection method for SST-5 and IMP datasets.
To this end, we use Amazon’s Mechanical Turk crowd-sourcing platform \cite{crowston2012amazon}.
Figure \ref{fig:app_sst5_interface} and \ref{fig:app_imip_interface} show the interfaces used to collect annotations from crowd workers for each task.
The top provides the summary, the middle provides detailed instructions, and then examples are shown. 
The whole task has 10 items per Human Interface Task (HIT). Workers were paid US\$1.0 per HIT on average, and all workers were paid for their work. 
To improve the quality of collected preference labels, we only hire the Master workers identified as high-performing workers from Amazon’s Mechanical Turk system. 
Overall, we gather at least 3 annotations for each sample. For the experiments with annotated samples, we use the same experimental setups with data pruning in Section\ref{sec5.1}.
Also, for the annotator disagreement, we report variance within the multiple annotations. We will release the annotated dataset for future research.

\section{Additional Results}\label{app_more_results}

\input{figures/acl23_figure12}
\input{figures/acl23_figure13}

\subsection{Data Pruning}
\input{tables/acl23_table8_info_dpp_ablation}
\input{figures/acl23_figure14}
\input{figures/acl23_figure15}

First, in Figure \ref{fig:app_fig4}, we plot the test accuracy of fine-tuned RoBERTa-large across different pruning ratios on CoLA, SST-2, RTE, and QNLI datasets; while the baseline methods suffer from inconsistent performance on different pruning ratio (for example, Hard and Ambig show good performance on low pruning ratio, but steeply degraded when the pruning ratio increases), \name{}-DPP shows the consistently outperforming performance in overall. 
In addition, we plot the dynamics during data pruning with \name{}-DPP in Figure \ref{fig:app_pruning_analysis}, similar to Figure \ref{fig:fig_confusion_pruning}. 
Here, one can observe that \name{}-DPP automatically finds the effective strategy adaptively. 
Finally, we present the ablation results of our component (1) \name{} and (2) DPP-based sampling method.
As shown in Table \ref{table:ablation_info_dpp}, the DPP-based sampling method provides a clear improvement in multidimensional space (vs Coreset). 
Furthermore, as \name{} provides a richer feature space than the standard classifier's embedding, such gain is further enlarged when they are combined.

\subsection{Active Learning}

In Figure \ref{fig:fig_app_al}, we present the test accuracy of fine-tuned BERT-base with each AL iteration on RTE and AGNEWS, respectively. 
Here, one can observe that \name{}-DPP shows comparable performance with the state-of-the-art baseline of AL.

\section{\name{} on Other Datasets}\label{app:info_other}

\input{figures/acl23_figure16}

In this section, we first present the correlation matrices between 23 meta-information on other datasets, similar to Figure \ref{fig:fig_confusion}. 
As one can see in Figure \ref{fig:app_conf_pruning}, the correlation matrices between different datasets (and tasks) are quite similar, which implies that the meta-information captures the general characteristic of datasets.
In addition, we present \name{} (bottom left and zoom in right) on other datasets (CoLA, WinoGrande, RTE, and SST-2) along with its classifier's embedding space (top left) and data map \cite{carbonera2015density} (middle left) in Figures \ref{fig:figure2_sst}, \ref{fig:figure2_rte}, \ref{fig:figure2_cola}, and \ref{fig:figure2_wino}.
Here, one can observe that \name{} successfully reveals distinctive regions again.

\input{figures/acl23_figure17}
\input{figures/acl23_figure18}
\input{figures/acl23_figure19}
\input{figures/acl23_figure20}

\section{Experiments to Verify Complementary Effect of Meta-information}\label{app_table2}
\input{tables/acl23_table9_ablation_multiple}

To further verify the complementary effect between multiple meta-information, we conduct simple toy experiments in this section. 
Similar to \cite{swayamdipta2020cartography}, we train a simple linear classifier on each feature space by assuming that gold labels of each task are available for training; for example, given sample is noisy-labeled or not. 
Here, we consider four different abnormality detection tasks with the QNLI dataset: mispredicted, mislabeled (or noisy labeled), out-of-distributed (OOD), and adversarial samples (Adv), respectively.    

In Table \ref{table:ablation}, one can verify that the accuracy increases as more meta-information is used; it implies that they are actually complementary and can provide richer information when they are jointly considered. 
In the end, \name{} shows a better performance than the original classifier's semantic embedding in all tested cases. 
Also, we consider the reduced feature space, $^{*}$\name{}, by applying the PCA-based feature selection method using the correlation matrix, and verify its comparable performance only using the half of meta-information. 
But, since only small costs are additionally required to use \name{} compared to $^{*}$\name{}, we use all 23 meta-information for our experiments. 
It is noteworthy that new meta-information can be easily included in our framework, and contribute to compose more informative feature space. In the remaining part of the section, we further provide the details of this experiment. 
Here, we use RoBERTa-large classifier \cite{liu2019roberta} fine-tuned on QNLI dataset \cite{wang2019glue}. 

1) Finding mispredicted samples: we train a single linear classifier with SGD optimizer on each feature space to classify whether a given sample is correctly predicted or not. Here, we assume that the binary labels that indicate whether a given sample is correctly predicted or not are available to train the linear classifier. Then, we only measure the performance on the test mispredicted samples. 

2) Detecting mis-labeled samples: following the setups in \cite{swayamdipta2020cartography}, we artificially impose the 10 \% label noise into training samples with high confidence (\textit{i.e.}, easy-to-learn). Then, we train a single linear classifier with SGD optimizer on each feature space to classify whether given samples has corrupted label or not. Here, we assume that the binary labels that indicate whether a given sample is corrupted or not are available to train the linear classifier. 

3) Detecting out-of-distribution samples: we consider the samples of QNLI's development set as inD and samples of MNLI's development set (both matched and mismatched) as OOD. Then, we train a single linear classifier with SGD optimizer on each feature space to classify whether the given sample is from inD or OOD. Here, we assume that the binary labels that indicate whether a given sample is inD or OOD are available to train the linear classifier. 

4) Detecting adversarial sentence: Here, we consider the task to classify whether the normal sentence is given (MNLI) or adversarially gathered sentence is given (ANLI \cite{nie2019adv_nli}). Then, we train a single linear classifier with SGD optimizer on each feature space to classify whether the given sample is normal or adversarial. Here, we assume that the binary labels that indicate whether a given sample is normal or adversarial are available to train the linear classifier. 

\section{Computational Cost\label{supp.cost}}
The computation cost of infoVerse depends on which meta-information is used for constructing it.
As introduced in Table \ref{summarization}, we considered meta-information with four categories (static measures, training dynamics, model uncertainty, and pre-trained knowledge).
The calculation of these categories requires (1, $E$, $T$, and 2) forward passes of the trained model for each sample, where $E$ denotes the total training epochs, and $T$ indicates the total number of trained models with different random seeds.
In the case of the proposed sampling method, it has $O(N^2M)$ time complexity when returning $N$ items from M total items, but we remark that it can be further reduced with a simple approximation \cite{chen2018fastdpp}.
Yet, we remark that our method does not require additional training costs since it only utilizes the trained models via standard way (cross entropy and multiple random seeds) and pre-trained models.

\jhk{For example, we measured the runtime of our approach using the CoLA dataset, with the time consumption in seconds: training/constructing infoVerse/DPP sampling consume 3852s/100s/10s, respectively.}
This demonstrates that our method's overall cost is relatively minor compared to training expenses. 
Moreover, it is worth noting that the cost of ours is only incurred once at initial construction.
It is also worth noting that all meta-information within the same category are obtained with the same forward passes, and there is no additional cost after infoVerse is constructed at once.

In addition, although this work did not focus much on reducing computational overhead, we believe simple practices can further reduce the cost of constructing infoVerse.
For example, substituting the additional forward passes by saving the outputs of models during training on the fly \cite{huang2017snapshot, tarvainen2017mean} or using a proxy of the original model with smaller architectures and fewer training epochs \cite{coleman2019selection}.

\end{document}

%% file: figures/acl23_figure1.tex
\begin{figure}[t!]
	\centering
	\includegraphics[width=0.85\columnwidth,clip]{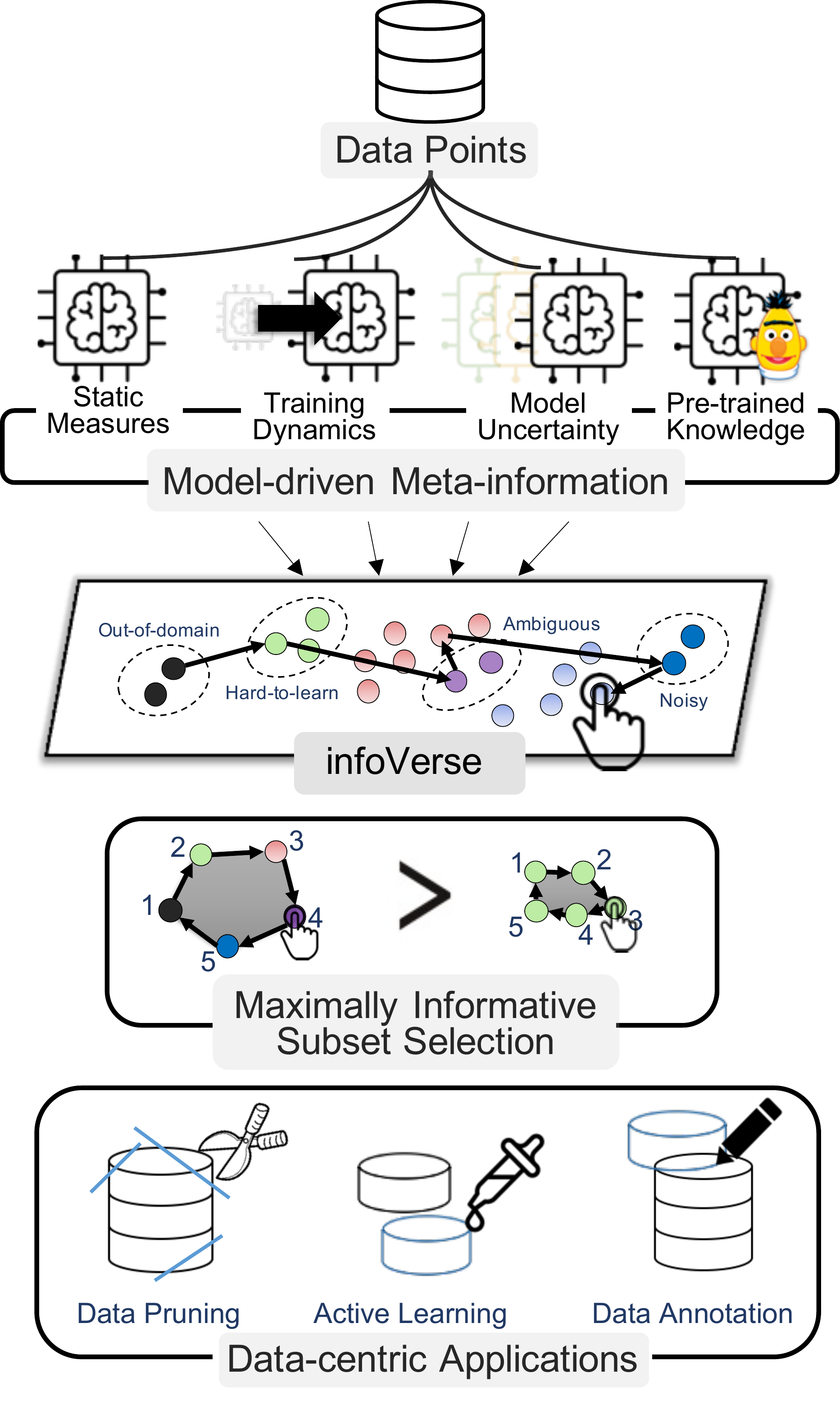}
    \vspace{-2mm}
    \caption{The proposed framework, \textbf{\name{}}. By incorporating multidimensional aspects of data characteristics, \name{} enables a better dataset characterization. By selecting maximally informative subsets on \name{}, we improve model performance on a variety of data-centric real-world problems like active learning.
    }
    \label{fig:figure1}
    \vspace{-3mm}
\end{figure}

%% file: tables/acl23_table1_features.tex
\begin{table}[t]
    \begin{center}
    \caption{Categorization of used meta-information.\vspace{-1mm}
    }
    \begin{adjustbox}{width=1.0\columnwidth}
    \begin{tabular}{@{}cll@{}}
        \toprule
        \multicolumn{2}{c}{\textbf{Categories}} & {\textbf{Meta-information}} \\
        \midrule 
        \multirow{2}{*}{\includegraphics[width=8mm,height=8mm]{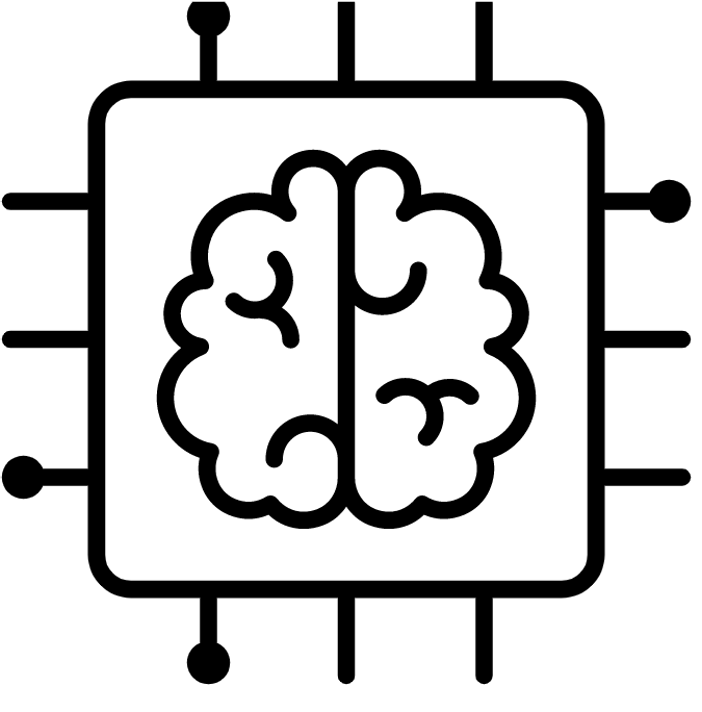}} & \textbf{Static} & Entropy, Confidence, BADGE   \\
        &\textbf{Measures} & Task Density, Relative Density score \\ \hdashline[0.4pt/2pt]
        \multirow{2}{*}{\includegraphics[width=12mm,height=8mm]{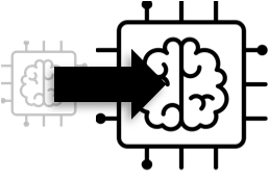}} & \textbf{Training} & Average Confidence, Variability  \\
        & \textbf{Dynamics} & Forgetting Number, Area Under Margin \\ \hdashline[0.4pt/2pt]
        \multirow{2}{*}{\includegraphics[width=12mm,height=8mm]{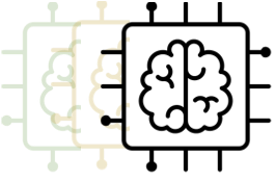}} & \textbf{Model} & \multirow{2}{*}{EL2N score, BALD, Variance Ratio}  \\
        & \textbf{Uncertainty} &     \\ \hdashline[0.4pt/2pt] 
        \multirow{2}{*}{\includegraphics[width=8mm,height=8mm]{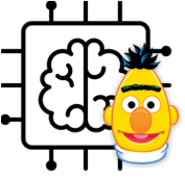}} & \textbf{Pre-trained} & \multirow{2}{*}{Sentence Density, Pseudo-log-likelihood}\\
        & \textbf{Knowledge} &  \\ 
        \bottomrule
    \end{tabular}
    \label{summarization} 
    \end{adjustbox}
    \vspace{-4mm}
    \end{center}
\end{table}

%% file: figures/acl23_figure2.tex
\begin{figure}[t]
	\centering
	\includegraphics[width=0.9\columnwidth, trim={5.8cm 1.5cm 1.5cm 0cm},clip]{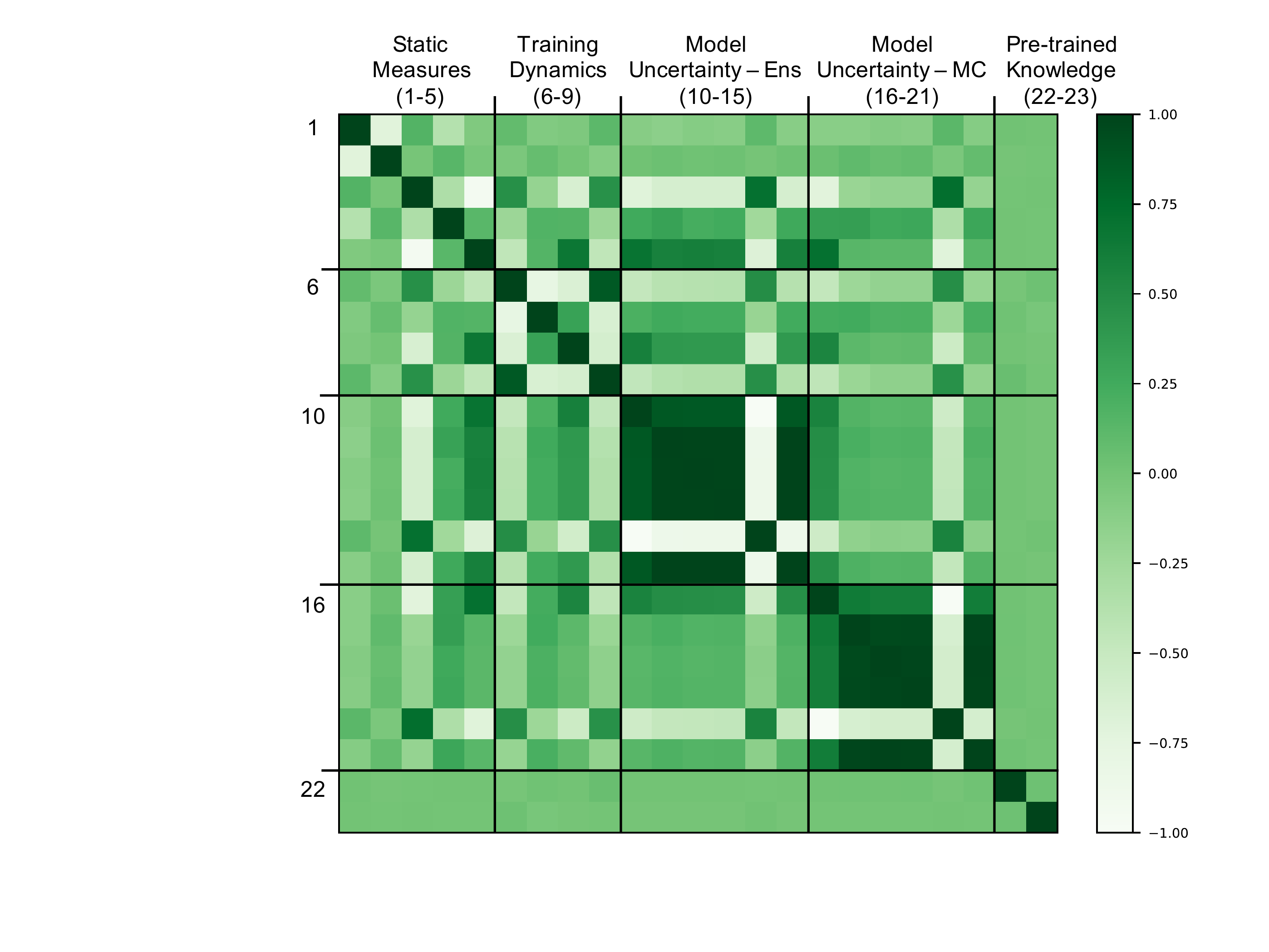}\vspace{-4.5mm}
    \caption{Correlation between meta-information considered in \name{} on QNLI dataset.\vspace{-4.5mm}}
    \label{fig:fig_confusion}
\end{figure}

%% file: figures/acl23_figure3.tex
\begin{figure}[t]
	\centering
	\includegraphics[width=1.0\columnwidth,trim=0.6cm 0.3cm 0.1cm 0.5cm, clip]{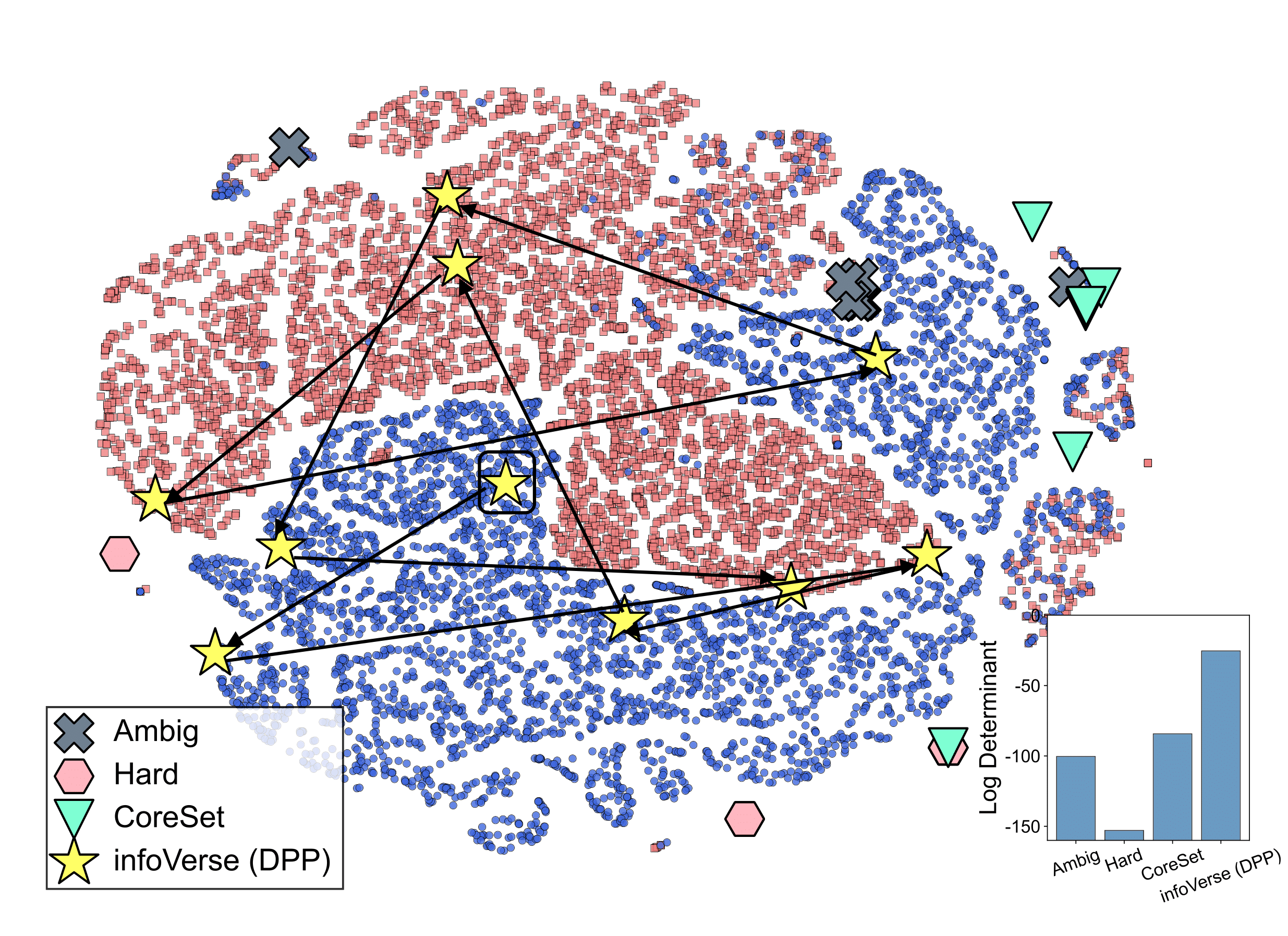}\vspace{-3mm}
    \caption{10 selected samples with different selection methods on QNLI dataset and their log determinant as a proxy measure for set-informativeness.\vspace{-3mm}}
    \label{fig:figure_dpp}
\end{figure}

%% file: figures/acl23_figure4.tex
\begin{figure*}[t]
	\centering
	\includegraphics[width=1.0\linewidth,trim={10mm 0 5mm 0},clip]{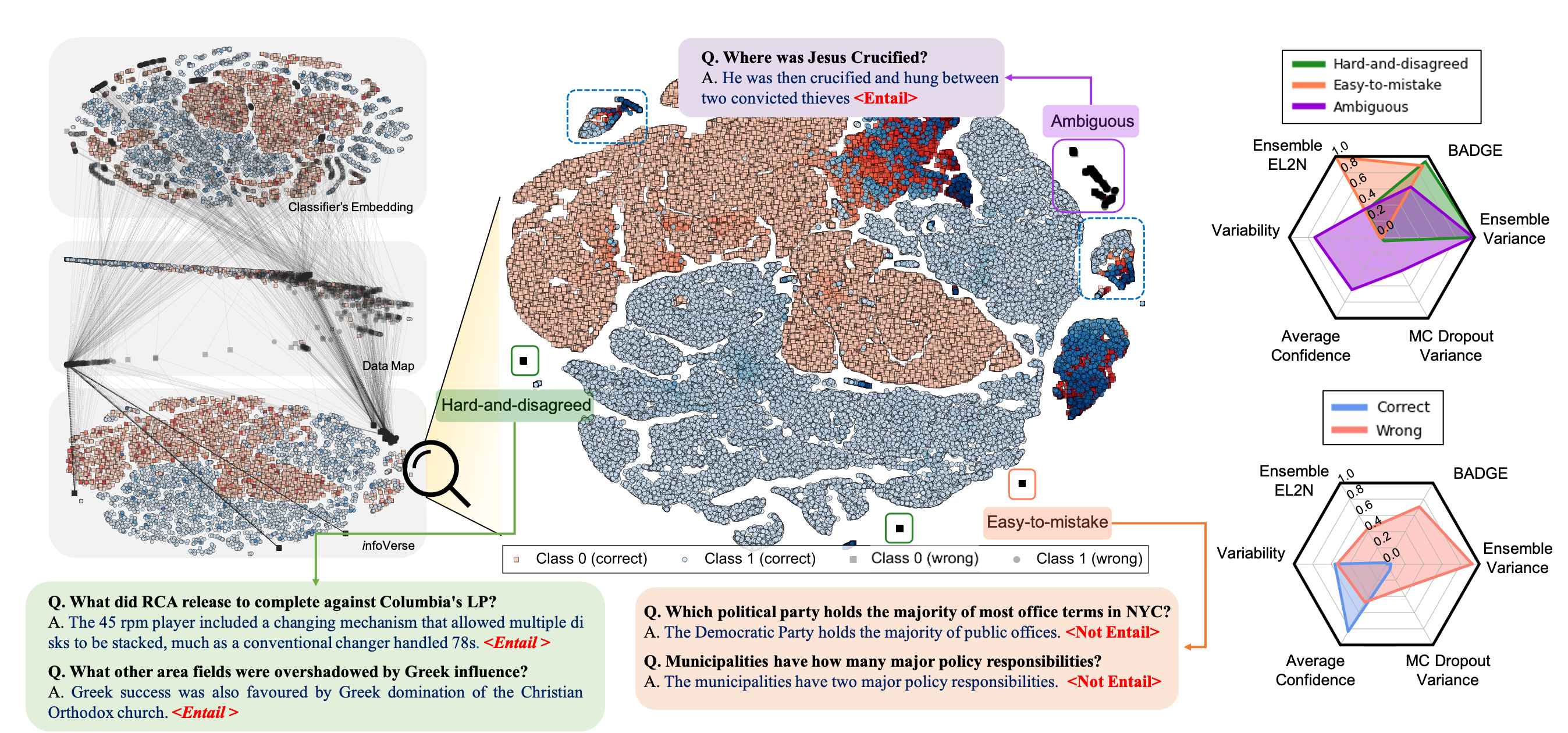}\vspace{-1mm}
    \caption{
    \name{} (bottom left) on QNLI dataset along with other feature spaces: classifier's embedding (top left) and \textit{data map} \cite{swayamdipta2020cartography} (middle left). (center) We present the zoomed version of \name{} with some examples of each distinctive region. We set high chroma to high variability samples for better interpretation. (right) Score distribution of each region characterized by \name{}.
    }\vspace{-3mm}
    \label{fig:figure2}
\end{figure*}

%% file: tables/acl23_table2_pruning.tex
\begin{table*}[t]
    \caption{Average test accuracy of fine-tuned RoBERTa-large over the eight different data pruning ratios.\vspace{-3mm}}
	\begin{center}
	\begin{adjustbox}{width=1.0\linewidth}
	\begin{tabular}{r|c|ccccccc|c}
 		\toprule
		Dataset & Random & Easy & Hard & Ambig & Forget & EL2N & Coreset & Dense & \name{}-DPP  \\ \midrule
		WinoGrande       & {73.1}\ms{0.09}  & {68.1}\ms{0.18}  & {69.6}\ms{0.31}  & {69.3}\ms{0.52}  & \underline{73.8}\ms{0.77}  & {72.2}\ms{1.02}  & {73.9}\ms{0.14}  & {72.9}\ms{0.31}  & \textbf{{74.6}}\ms{0.24}  \\ 
		CoLA       & {60.7}\ms{0.63}  & {32.2}\ms{0.94}  & {41.1}\ms{0.16}  & {41.0}\ms{0.73}  & {59.1}\ms{0.54}  & {47.6}\ms{1.07}  & {55.5}\ms{0.93}  & \underline{61.2}\ms{0.43}  & \textbf{{62.5}}\ms{0.14}  \\
		RTE        & {76.7}\ms{0.75}  & {73.5}\ms{0.21}  & {56.2}\ms{1.59}  & {56.3}\ms{2.26}  & {71.0}\ms{0.93}  & {61.4}\ms{0.81}  & {56.3}\ms{0.31}  & \underline{78.1}\ms{1.20}  & \textbf{{78.5}}\ms{0.60} \\
		QNLI       & \underline{92.9}\ms{0.25}  & {70.0}\ms{0.07}  & {79.0}\ms{0.34}  & {80.0}\ms{0.49}  & {92.2}\ms{0.31}  & {82.8}\ms{1.85}  & {84.2}\ms{0.28}  & {92.1}\ms{0.30}  & \textbf{{93.1}}\ms{0.17}  \\ 
		SST-2      & \underline{95.3}\ms{0.10}  & {65.6}\ms{0.64}  & {88.7}\ms{0.29}  & {93.1}\ms{0.57}  & {95.2}\ms{0.09}  & {90.5}\ms{0.39}  & {92.9}\ms{0.48}  & {94.4}\ms{0.07}  & \textbf{{95.7}}\ms{0.10}  \\ 
		\bottomrule
	\end{tabular}
    \end{adjustbox}
    \end{center}
    \label{table:pruning}
    \vspace{-5mm}
\end{table*}

%% file: figures/acl23_figure5.tex
\begin{figure}[t!]
\begin{center}
    {
        \includegraphics[width=0.45\textwidth]{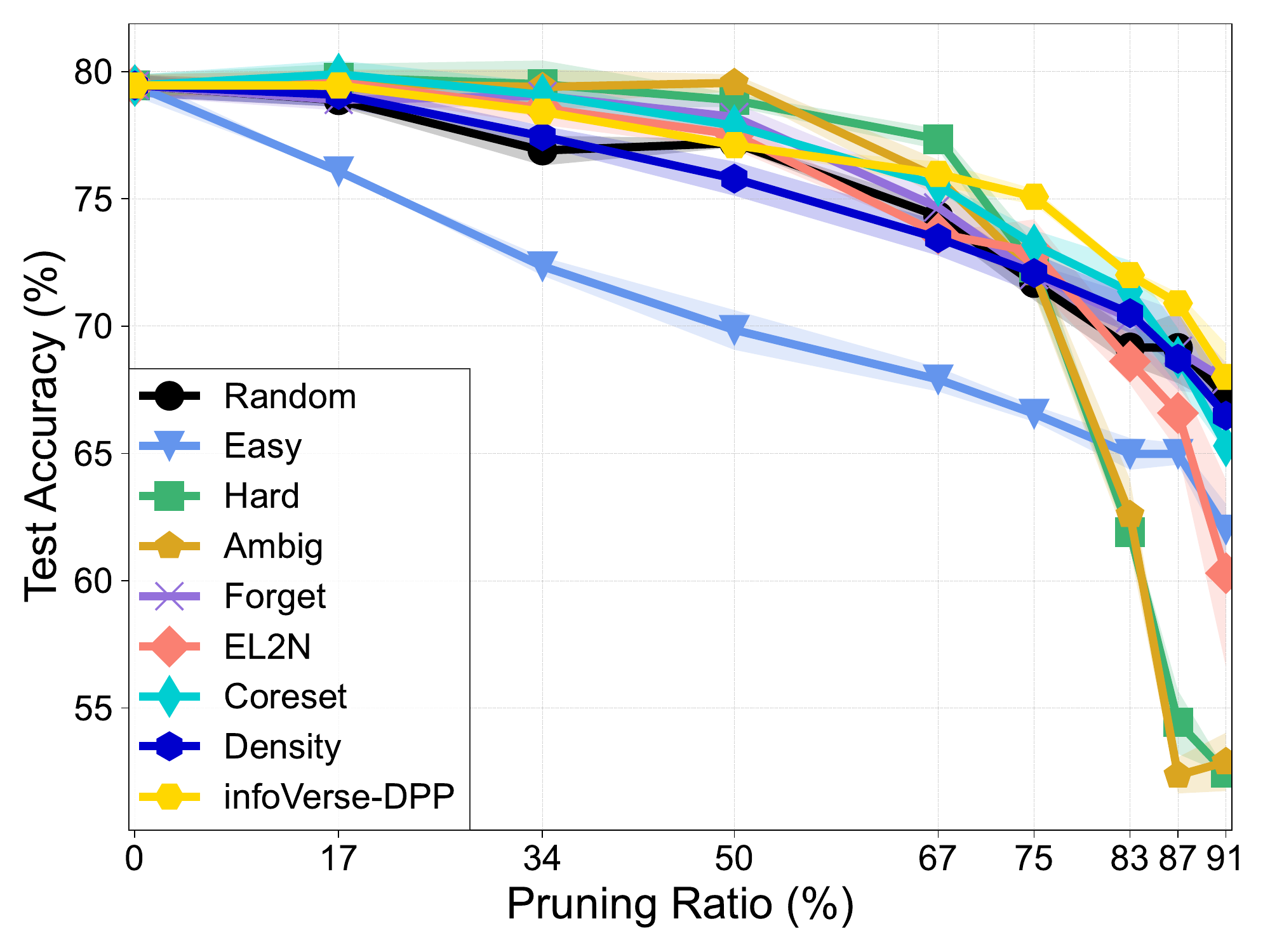}
    \vspace{-5mm}
    }
\end{center}
\caption{Test accuracy of RoBERTa-large fine-tuned on the pruned training dataset of WinoGrande. \vspace{-3mm}}
\label{fig:fig_main_pruning}
\end{figure}

%% file: tables/acl23_table3_pruning_ablation.tex
\begin{table}[t]
	\begin{center}
	\caption{Ablation study about the effectiveness of each category of meta-information with \name{}-DPP.}
	\vspace{-0.07in}
	\label{table:ablation_info_feature}
    \begin{adjustbox}{width=0.85\linewidth}
	\begin{tabular}{lcc}
		\toprule
		Categories  & WinoGrande & CoLA
		              \\ \midrule
		Static Measures & {72.7}\ms{0.24} & {60.2}\ms{0.19} \\
		Training Dynamics & {73.5}\ms{0.35} & {62.2}\ms{0.41} \\
		Model Uncertainty  & {71.5}\ms{1.47} & {60.6}\ms{0.36} \\
		MC-Model Uncertainty & {70.2}\ms{0.80} & {58.9}\ms{0.82} \\
		Pre-trained Knowledge & {72.8}\ms{0.84} & {56.3}\ms{0.51} \\ \midrule
		\name{} & \textbf{74.6}\ms{0.24} & \textbf{62.5}\ms{0.14} \\
		\bottomrule
    \vspace{-0.1in}
	\end{tabular}
    \end{adjustbox}
    \end{center}
    \vspace{-4mm}
\end{table}

%% file: figures/acl23_figure6.tex
\begin{figure}[t]
	\centering
	\vspace{-6mm}
	\includegraphics[width=1.0\columnwidth, trim={1.0cm 2.5cm 0.5cm 1.5cm},clip]{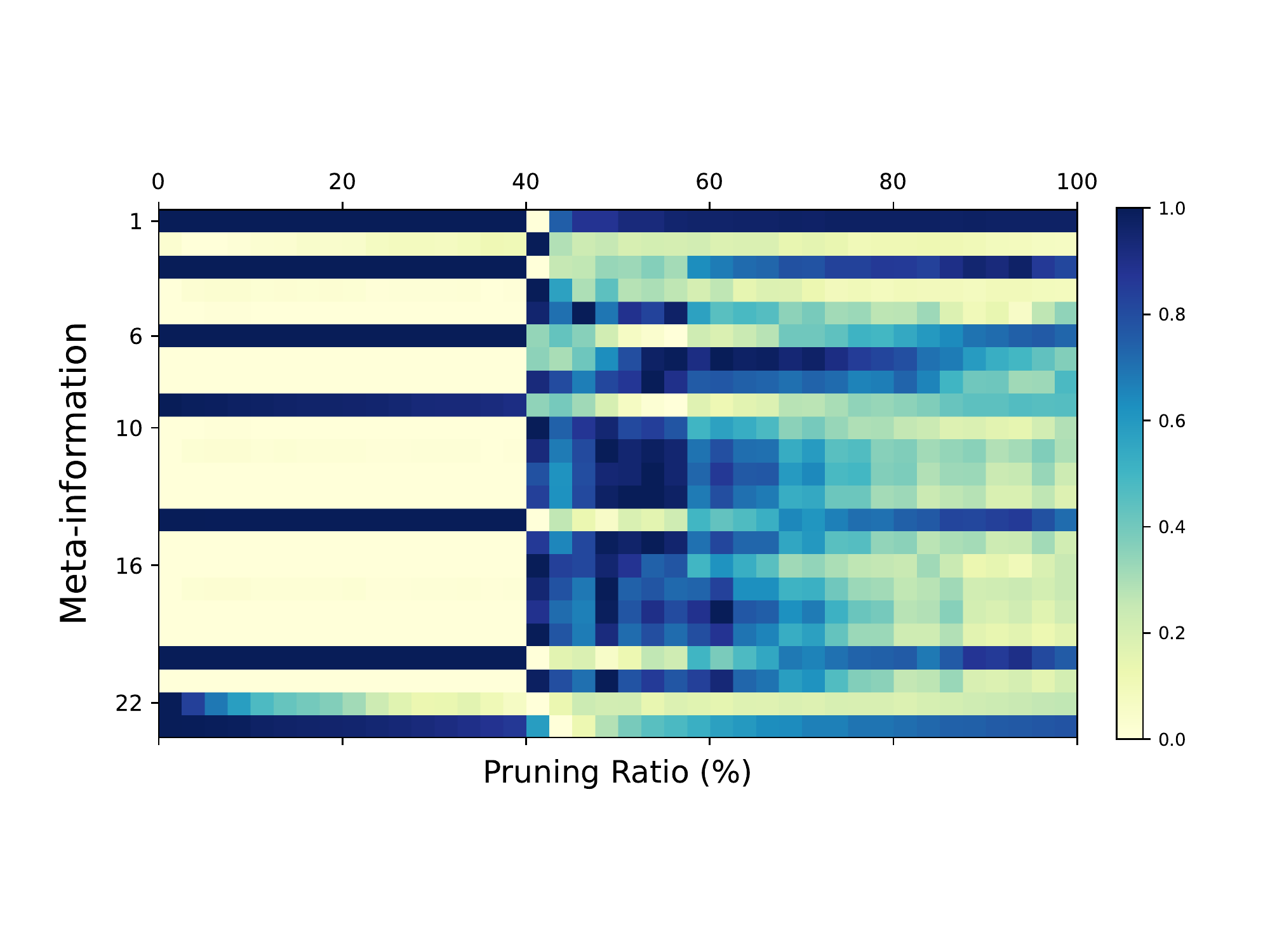}
	\vspace{-5mm}
    \caption{Majority measurement of the selected samples dynamically change as the pruning ratio increases (\name{}-DPP on SST-2):
    \textit{confidence} (0-40\%) $\rightarrow$ \textit{model uncertainty} (40-60\%) $\rightarrow$ \textit{variability} (60-85\%). \vspace{-3mm}}
    \label{fig:fig_confusion_pruning}
\end{figure}

%% file: tables/acl23_table4_active.tex
\begin{table*}[ht]
    \caption{Average test accuracy of fine-tuned BERT-base over the nine AL iterations.\vspace{-3mm}}
	\begin{center}
	\begin{adjustbox}{width=1.0\linewidth}
	\begin{tabular}{r|c| cccccc|c}
 		\toprule
		Dataset & Random & Entropy & BALD & BERT-KM & FT-BERT-KM & BADGE & ALPS & \name{}-DPP  \\ \midrule
		AGNEWS        & {89.9}\ms{0.25}  &  {90.6}\ms{0.17}  & {\textbf{90.8}}\ms{0.32}  & {89.8}\ms{0.34}  & {90.7}\ms{0.18}  & {90.6}\ms{0.21}  & {89.3}\ms{0.32}  & {90.6}\ms{0.19} \\ 
		SST-2        & {88.8}\ms{0.62}  &  {89.9}\ms{0.81}  & {89.3}\ms{0.63}  & {89.3}\ms{0.57}  & {89.4}\ms{0.64}  & {89.8}\ms{0.63}   & {89.2}\ms{0.66}  & {\textbf{90.3}}\ms{0.57}  \\
		RTE        & {60.9}\ms{2.80}  &  {60.5}\ms{1.92}  & {60.9}\ms{2.20}  & {60.7}\ms{0.06}  & {59.6}\ms{2.39}  & {59.8}\ms{2.29}  & {58.3}\ms{3.00}  & {\textbf{61.5}}\ms{2.46}
		\\ \bottomrule
	\end{tabular}
    \end{adjustbox}
    \end{center}
    \label{table:al}
    \vspace{-4mm}
\end{table*}

%% file: figures/acl23_figure7.tex
\begin{figure}[t]
\begin{center}
    {
    \includegraphics[width=0.45 \textwidth]{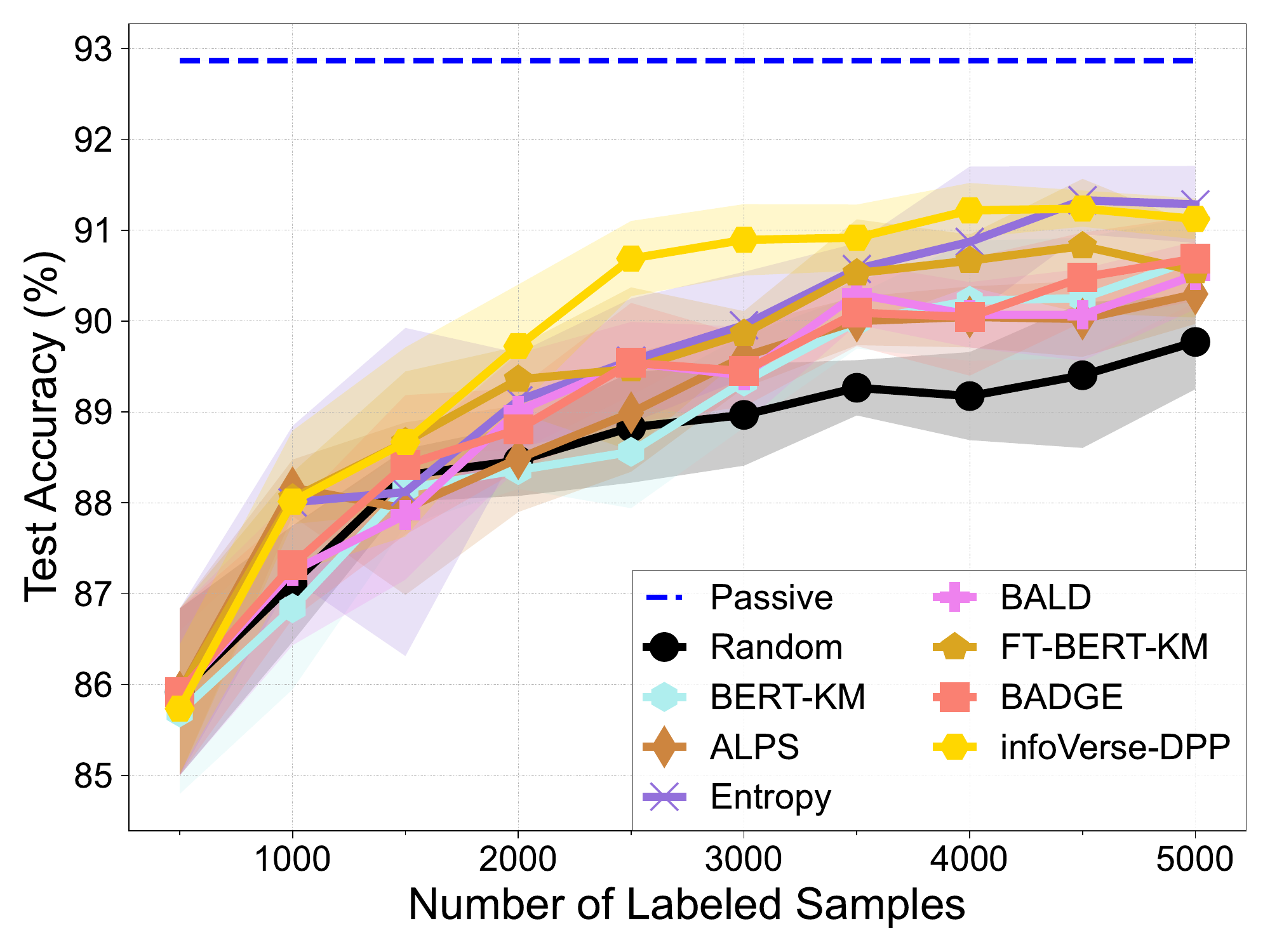}
    \label{fig:fig4a}
    \vspace{-3mm}
    }
\end{center}
\caption{Test accuracy of BERT-base fine-tuned on the labeled samples by each AL method on SST-2. 
\vspace{-3mm}}
\label{fig:fig_app}
\end{figure}

%% file: figures/acl23_figure8.tex
\begin{figure*}[t]
\begin{center}
    {
    \subfigure[Average Confidence]
        {
        \includegraphics[width=0.24\textwidth]{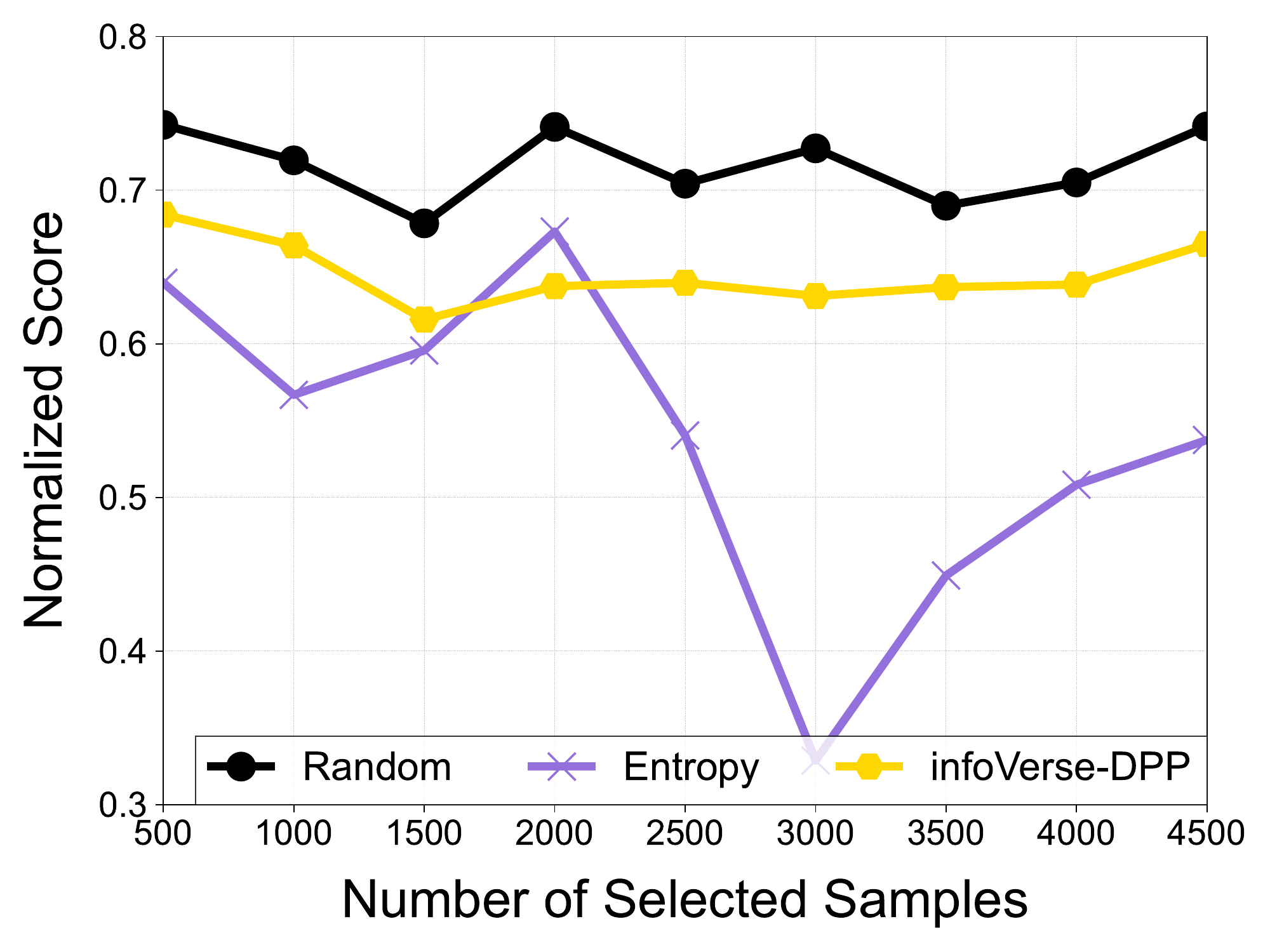}
        \label{fig:1a}
        }
        \hspace{-3mm}
    \subfigure[Variability]
        {
        \includegraphics[width=0.24\textwidth]{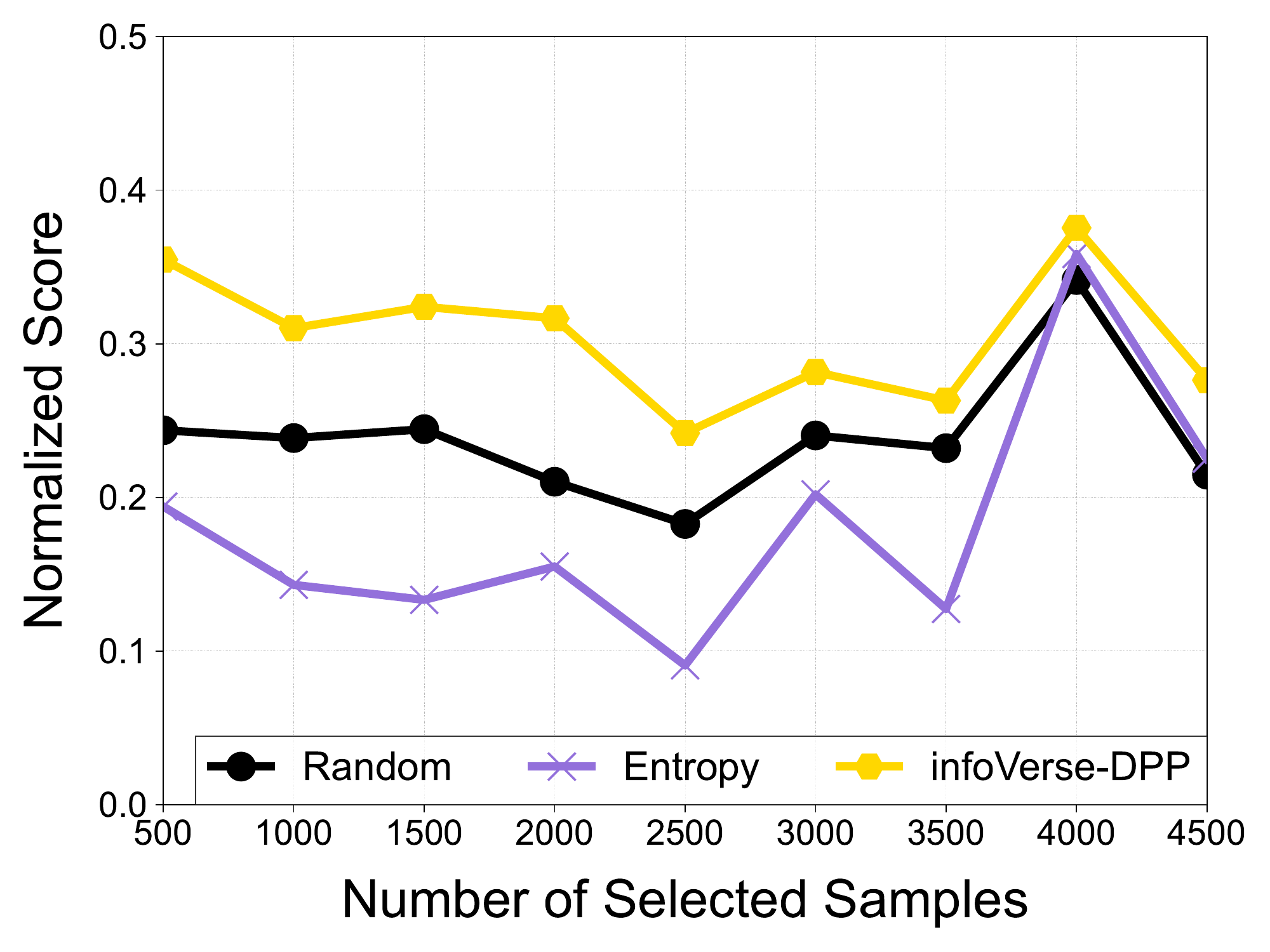}
        \label{fig:1b}
        }
        \hspace{-3mm}
    \subfigure[Ensemble Entropy]
        {
        \includegraphics[width=0.24\textwidth]{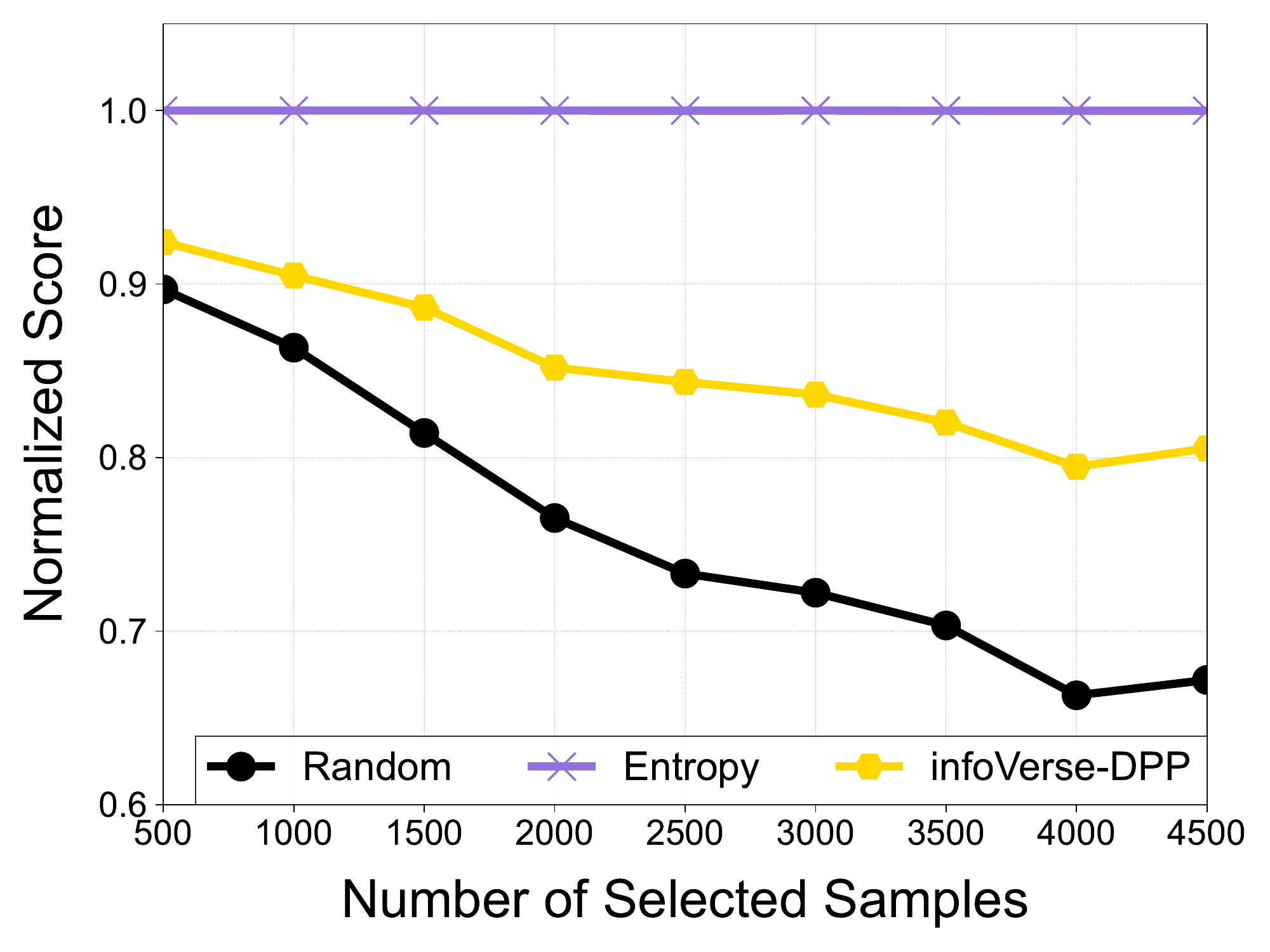}
        \label{fig:1c}
        } 
        \hspace{-3mm}
    \subfigure[Sentence Density]
        {
        \includegraphics[width=0.24\textwidth]{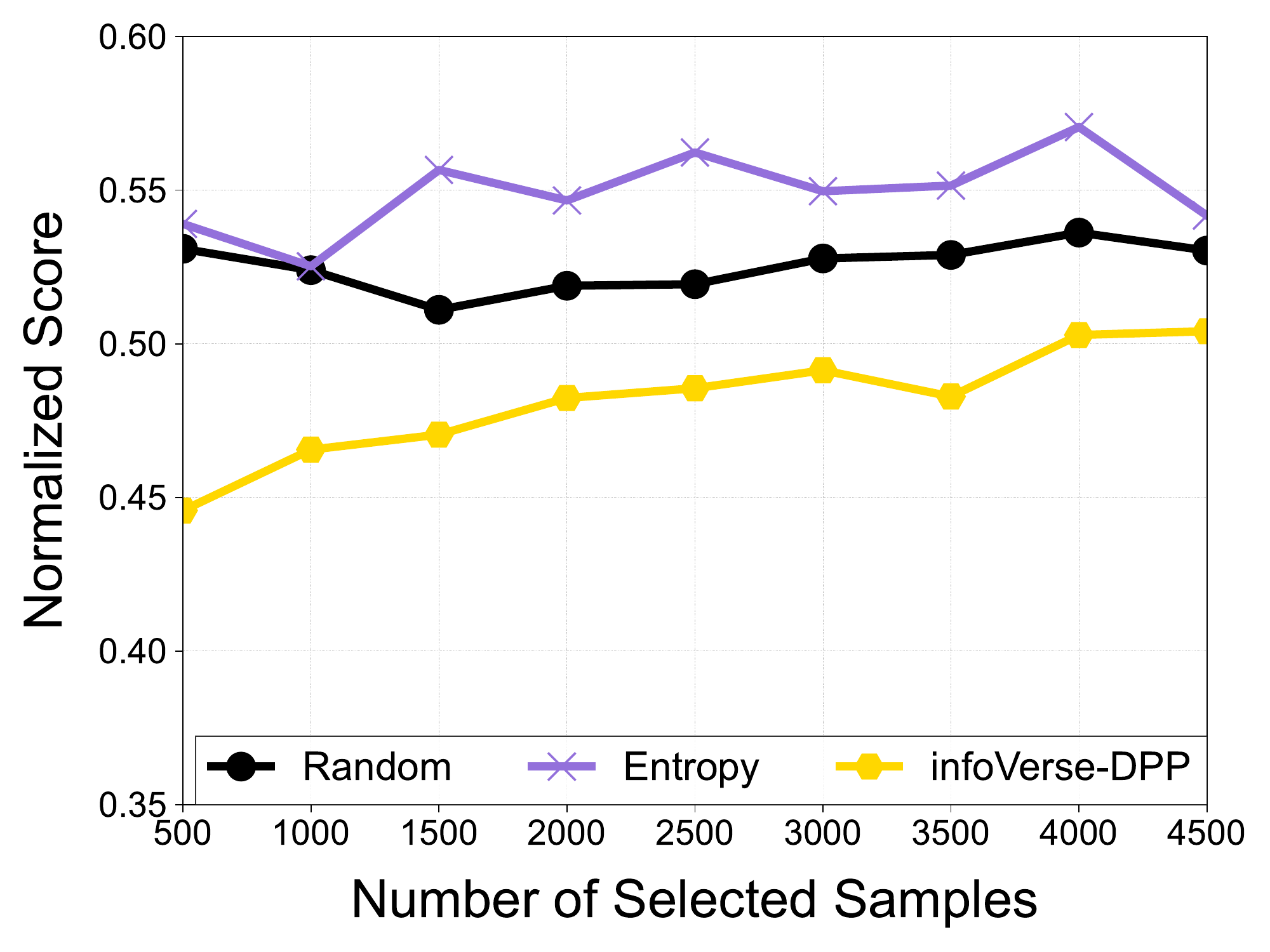}
        \label{fig:1d}
        } 
    }
\end{center}
\vspace{-0.1in}
\caption{Comparison of the selected samples with different AL methods (\textit{Random}, \textit{Entropy}, and \textit{Ours}) on SST-2. 
}
\vspace{-0.1in}
\label{fig:active_analysis}
\end{figure*}

%% file: figures/acl23_figure9.tex
\begin{figure}[t!]
	\centering
	\vspace{-2mm}
	\includegraphics[width=0.7\columnwidth]{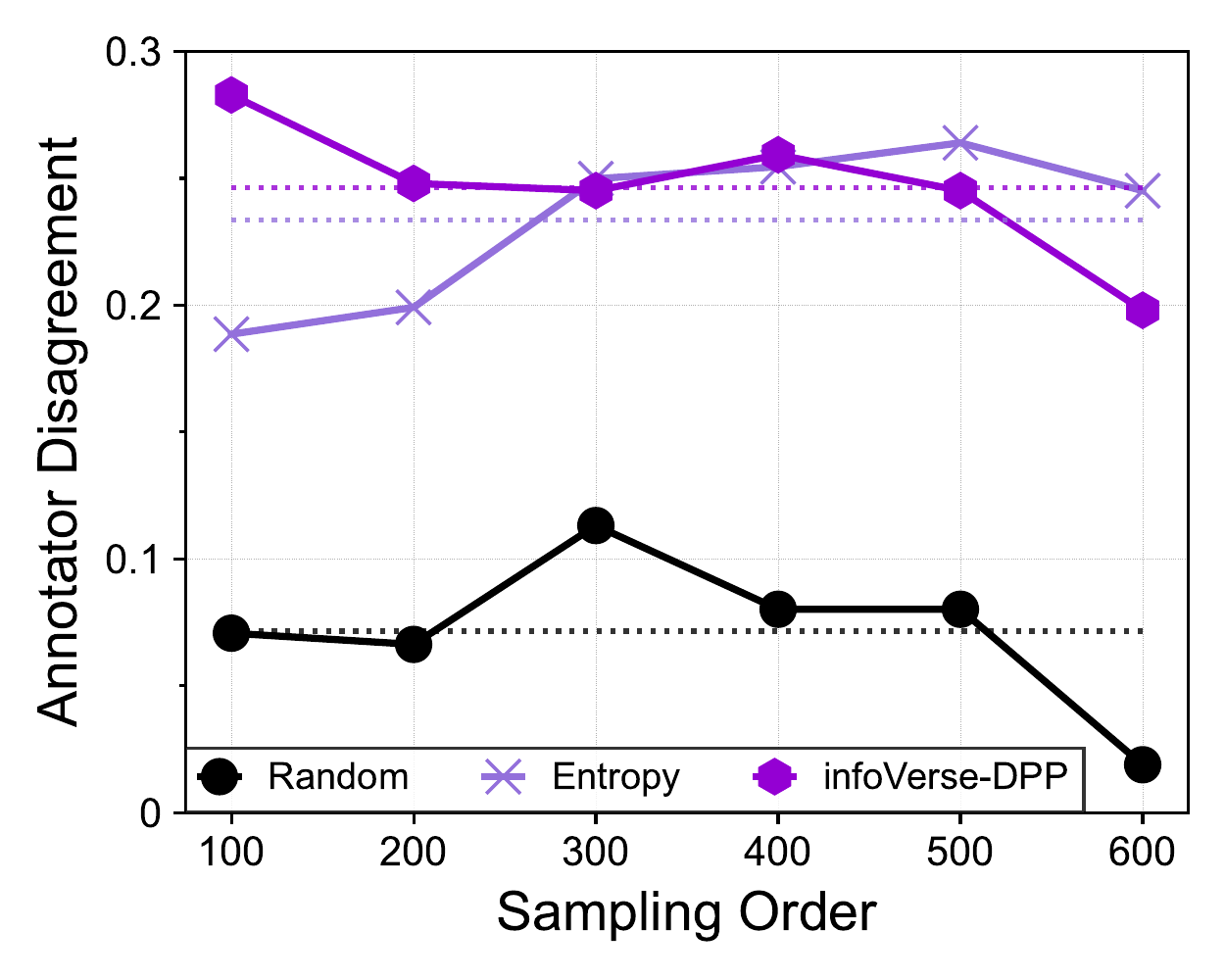}
	\vspace{-2mm}
    \caption{Annotator disagreement on IMP.}
    \vspace{-2mm}
    \label{fig:figure_disagree}
\end{figure}

%% file: tables/acl23_table5_real_annotation.tex
\begin{table}[t]
    \caption{Test accuracy of RoBERTa-large fine-tuend on the annotated dataset together with the original training dataset. 1,000 samples for SST-5 and 600 samples for IMP are additionally annotated, respectively. }\vspace{-3mm}
	\begin{center}
	\begin{adjustbox}{width=1.0\columnwidth}
	\begin{tabular}{r|cccc}
 		\toprule
		Dataset  & Original & Random & Entropy & infoVerse-DPP \\ \midrule
		SST-5    & {58.2}\ms{0.66} & {58.2}\ms{0.72} & {58.4}\ms{0.83} & \textbf{58.8}\ms{0.95} \\
		IMP      & {88.6}\ms{0.67} & {88.9}\ms{0.40} & {88.8}\ms{0.78} & \textbf{89.0}\ms{0.50} \\ 
 		\bottomrule
	\end{tabular}
    \end{adjustbox}
    \end{center}
    \label{table:data_annotation}
    \vspace{-5mm}
\end{table}

%% file: tables/acl23_table6_features_details.tex
\begin{table*}[t]
    \begin{center}
    \caption{Categorization of used meta-information to construct \name{}. The arrow between the parentheses indicates more informative direction for each measure: e.g., less confident data $(\downarrow)$ are less likely seen so more informative.\vspace{-2mm}
    }
    \begin{adjustbox}{width=0.95\linewidth}
    \begin{tabular}{@{}clcc@{}}
        \textbf{Categories} & \multicolumn{2}{c}{\textbf{Meta-information}} & \textbf{Description} \\
        \midrule 
        \multirow{6}{*}{\makecell{Static \\ Measures}}   & \textit{Confidence} & $(\downarrow)$  & \makecell[l]{$\bullet$ Predictive probability to true label}  \\
                                                                 & \textit{Entropy} & $(\uparrow)$     & \makecell[l]{$\bullet$ Entropy of the predictive probability}  \\
                                                                 & \textit{BADGE}  & $(\uparrow)$      & \makecell[l]{$\bullet$ Norm of the gradient with respect to parameters in the final (linear) layer}  \\
                                                                 & \textit{Task Density} & $(\downarrow)$     & \makecell[l]{$\bullet$ Euclidean distance to the $K_{th}$ nearest element on the contextualized embedding \\ from fine-tuned classifier}  \\
                                                                 & \textit{Relative Density} & $(\downarrow)$     & \makecell[l]{$\bullet$ Difference of \textit{Task density} within other class' samples and true class samples.} \\
        \midrule
        \multirow{6}{*}{\makecell{Training \\ Dynamics}}   & \textit{Confidence} & $(\downarrow)$        & \makecell[l]{$\bullet$ Average predictive probability to true label over the training epochs}  \\
                                                           & \textit{Variability} & $(\uparrow)$         & \makecell[l]{$\bullet$ Variance of predictive probability to true label over the training epochs}  \\
                                                           & \textit{Forgetting number} & $(\uparrow)$   & \makecell[l]{$\bullet$ Summation of \textit{forgetting} over the training epochs: Sample $i$ undergoes \textit{forgetting} \\ when accuracy of $i$ decreases between two consecutive epochs}  \\
                                                           & \textit{Area Under Margin} & $(\downarrow)$ & \makecell[l]{$\bullet$ Average \textit{margin} over the training epochs: \textit{Margin} captures how much larger \\ the assigned logit is than all other logits} \\
        \midrule
        \multirow{6}{*}{\makecell{Model \\ Uncertainty \\ (Ens or MC)}}   & \textit{EL2N score} & $(\uparrow)$      & \makecell[l]{$\bullet$ Approximation for the gradient norm of expected loss which bounds the expected \\ change in loss for sample $i$ caused by removing it from training} \\
                                                                  & \textit{Entropy} & $(\uparrow)$         & \makecell[l]{$\bullet$ Entropy of predictive probability over multiple models with different random seeds}   \\
                                                                  & \textit{BALD} & $(\uparrow)$            & \makecell[l]{$\bullet$ Mutual information between data-points and model's weights}  \\
                                                                  & \textit{Variation Ratio} & $(\uparrow)$  & \makecell[l]{$\bullet$ Proportion of predicted labels that are not coincided with the average prediction} \\
                                                                  & \textit{Confidence} & $(\downarrow)$        & \makecell[l]{$\bullet$ Average predictive probability to true label between the models}  \\
                                                           & \textit{Variability} & $(\uparrow)$         & \makecell[l]{$\bullet$ Variance of predictive probability to true label between the models}  \\
        \midrule
        \multirow{3}{*}{\makecell{Pre-trained \\ Knowledge}}       & \textit{PLL} & $(\uparrow)$           & \makecell[l]{$\bullet$ Pseudo-Log-Likelihood (PLL) score from pre-trained Masked Language Models}   \\
                                                                  & \textit{Semantical Density} & $(\downarrow)$  & \makecell[l]{$\bullet$ Euclidean distance to the $K_{th}$ nearest element on the contextualized embedding \\ from pre-trained sentence encoder, e.g., sBERT} \\
        \bottomrule
    \end{tabular}
    \vspace{-4mm}
    \label{citation-guide} 
    \end{adjustbox}
    \end{center}
\end{table*}

%% file: tables/acl23_table7_datasets.tex
\begin{table*}[ht]
    \centering
    \caption{Dataset statistics used in experiments}
    \label{tab:dataset}
    \begin{tabular}{lcccl}
    \toprule
    Dataset & Language & Domain & Classes & Train / Dev  \\\hline
    QNLI & EN & Natural Language Inference  & 2 & 104k / 5.4k \\
    SST-2 & EN & Sentiment Analysis & 2 & 67k / 873 \\
    WinoGrande & EN & Commonsense Reasoning & 2 & 40k / 1.2k \\
    CoLA & EN & Linguistic Acceptability & 2 & 8.5k / 1.0k \\
    RTE & EN & Natural Language Inference & 2 & 2.5k / 278 \\
    AGNEWS & EN & Topic Classification & 4 & 110K / 7.6k \\

    \bottomrule
    \end{tabular}
\end{table*}

%% file: figures/acl23_figure10.tex
\begin{figure*}[t]
	\centering
	\includegraphics[width=0.9\linewidth]{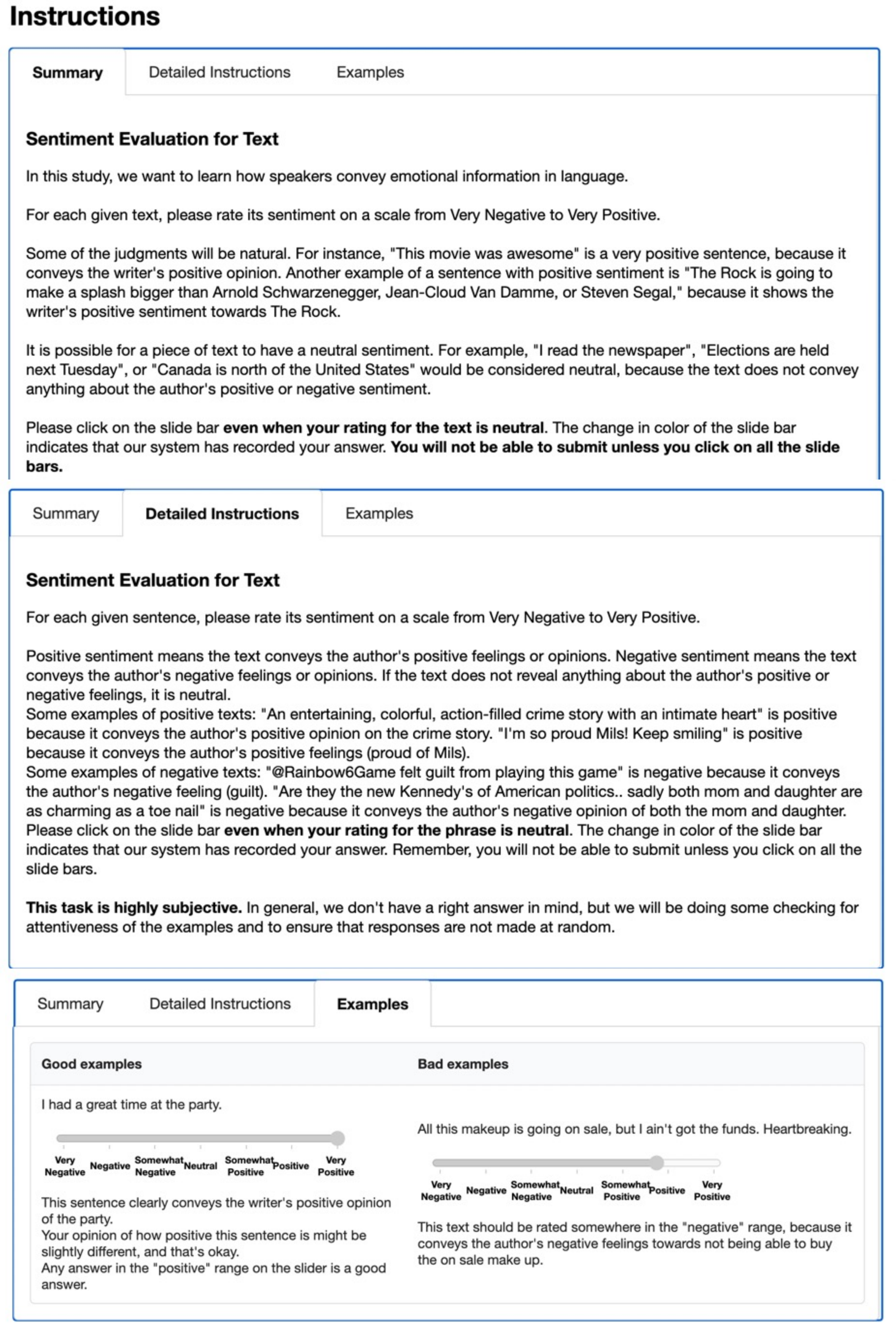}
    \caption{Interface to collect annotation from crowd workers for sentiment classification for SST-5 }\vspace{-3mm}
    \label{fig:app_sst5_interface}
\end{figure*}

%% file: figures/acl23_figure11.tex
\begin{figure*}[t]
	\centering
	\includegraphics[width=0.9\linewidth]{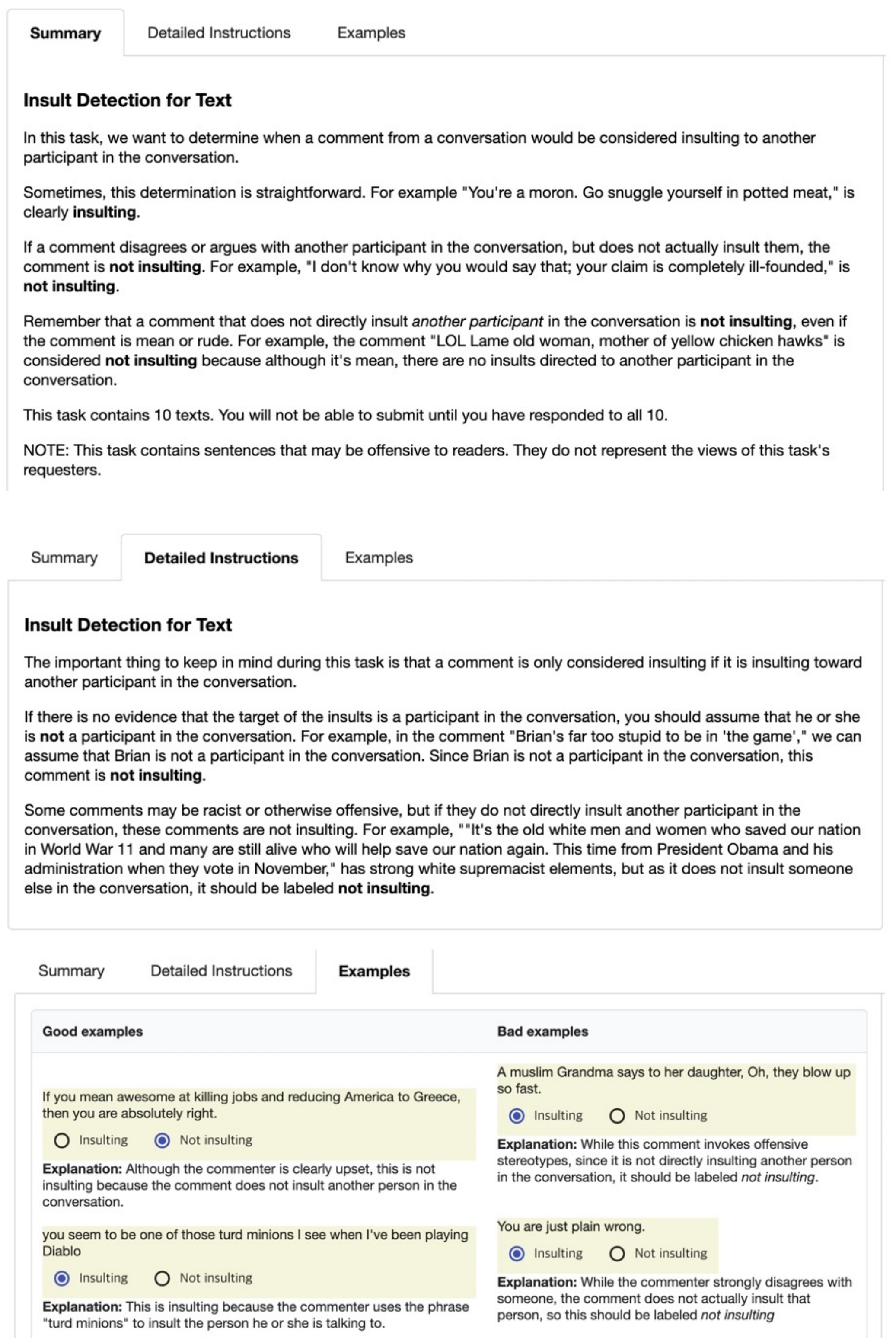}
    \caption{Interface to collect annotation from crowd workers for insult detection for IMP dataset.}\vspace{-3mm}
    \label{fig:app_imip_interface}
\end{figure*}

%% file: figures/acl23_figure12.tex
\begin{figure}[t]
\begin{center}
    {
    \subfigure[RTE]
        {
        \includegraphics[width=0.43\textwidth]{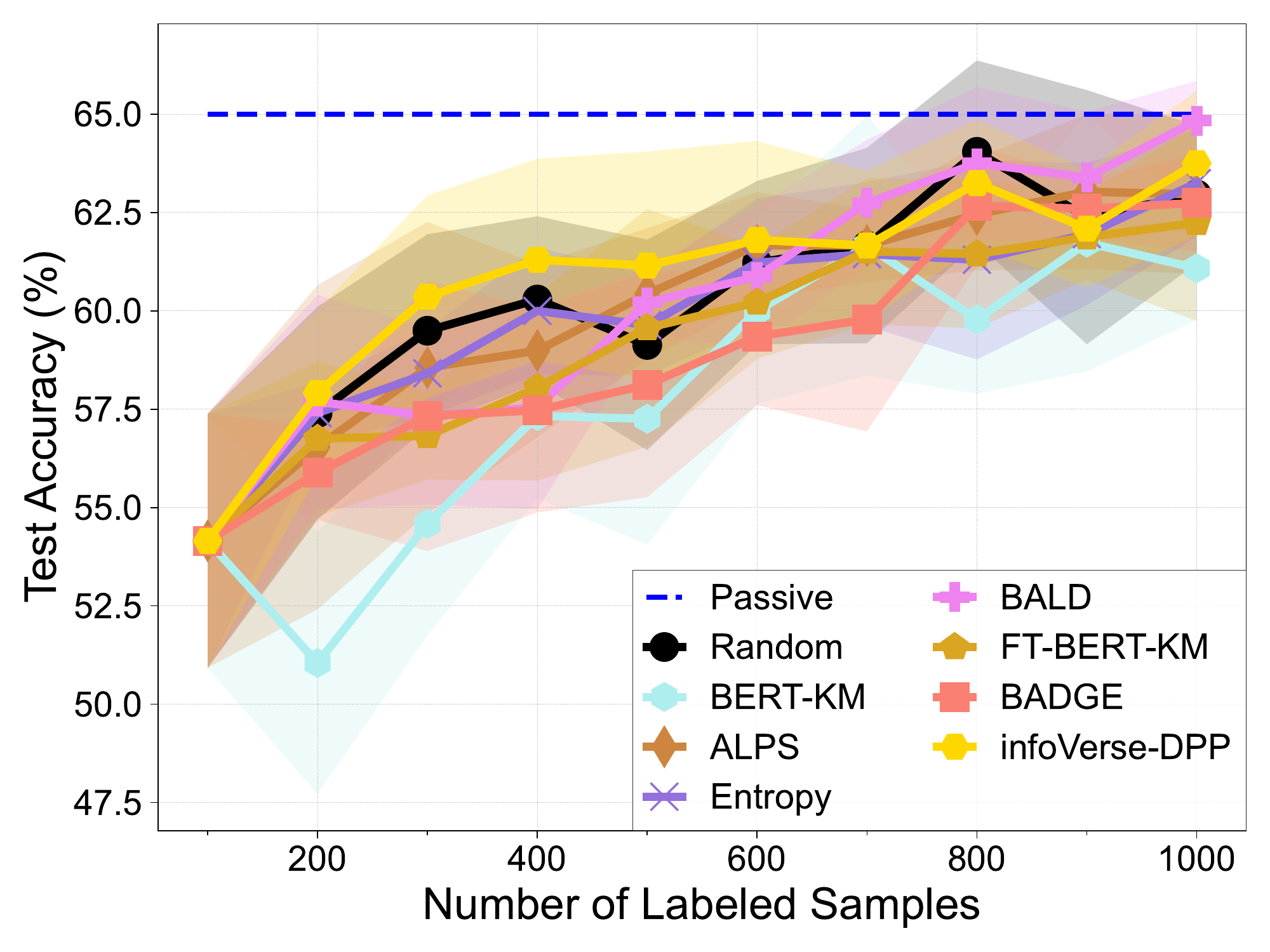}
        \label{fig:fig3b}
        } 
    \subfigure[AGNEWS]
        {
        \includegraphics[width=0.43\textwidth]{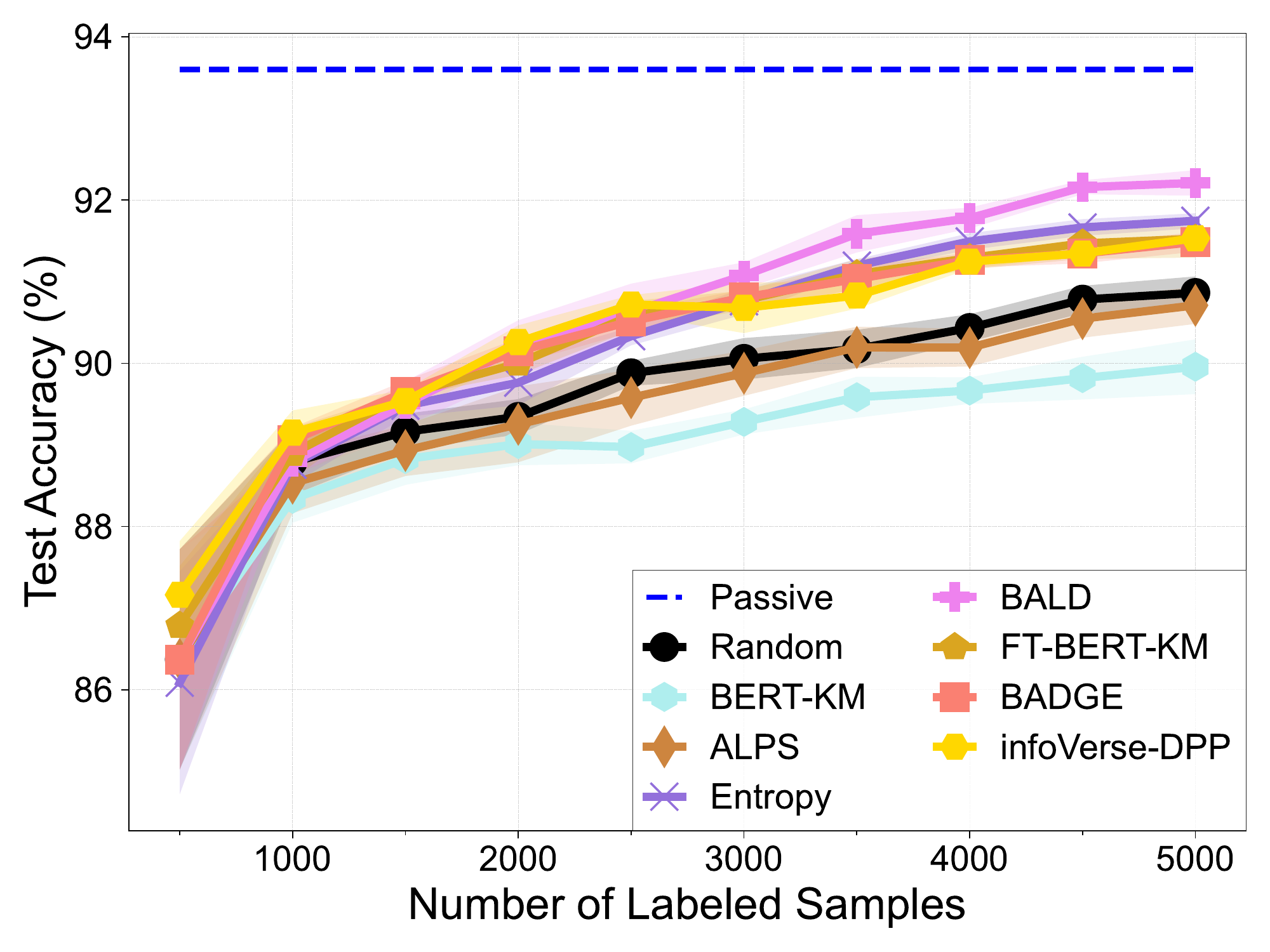}
        \label{fig:fig3c}
        }
    }
\end{center}
\caption{Test accuracy of BERT-base classifier fine-tuned using the labeled samples by each AL method on (a) RTE and (b) AGNEWS.}
\label{fig:fig_app_al}
\end{figure}

%% file: figures/acl23_figure13.tex
\begin{figure}[t]
\begin{center}
{
    \subfigure[CoLA]
    {
        \includegraphics[width=0.43\textwidth]{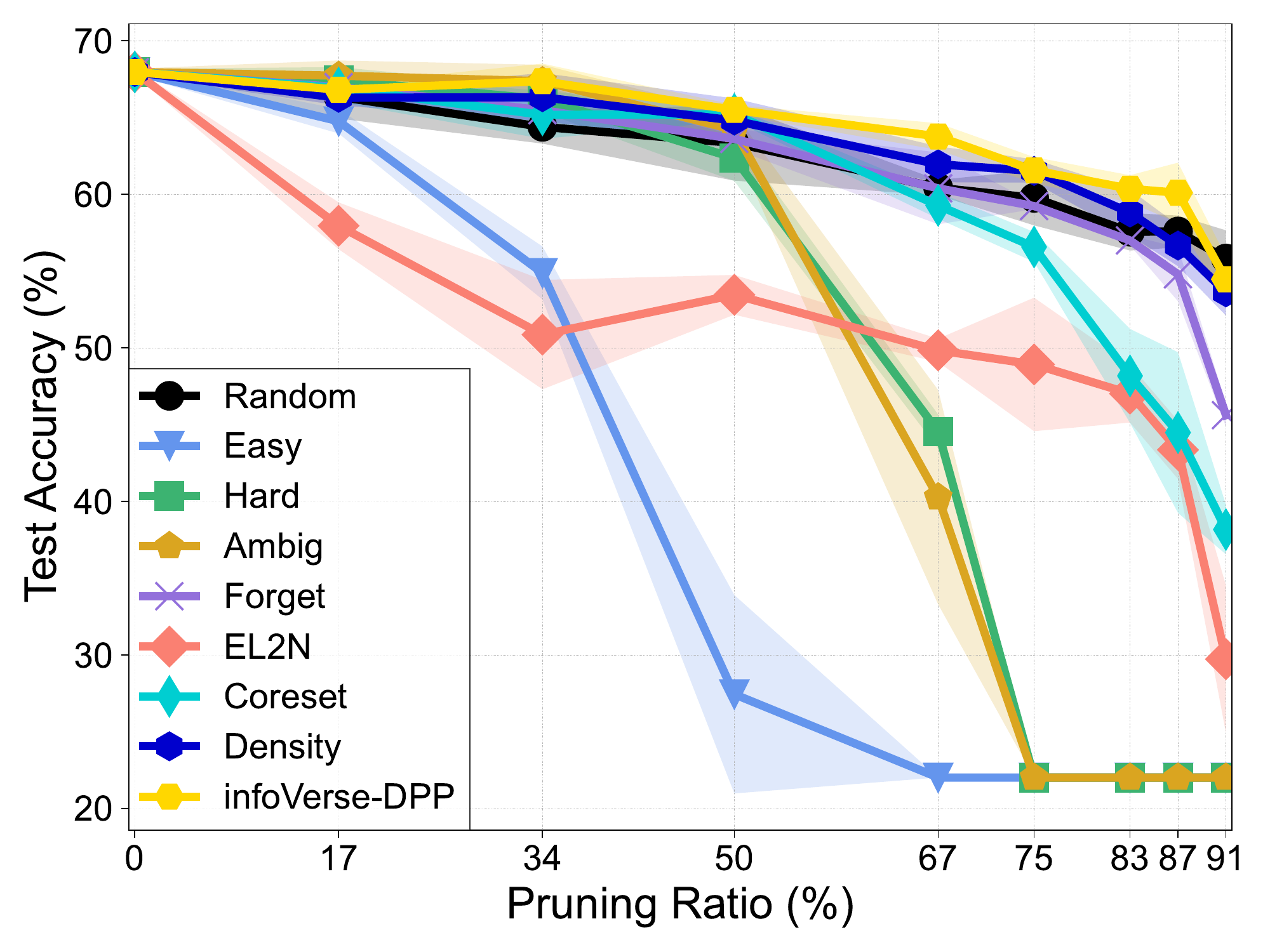}
        \label{fig:app_ds2}
    }
    \vspace{-2mm}
    \subfigure[SST-2]
    {
        \includegraphics[width=0.43\textwidth]{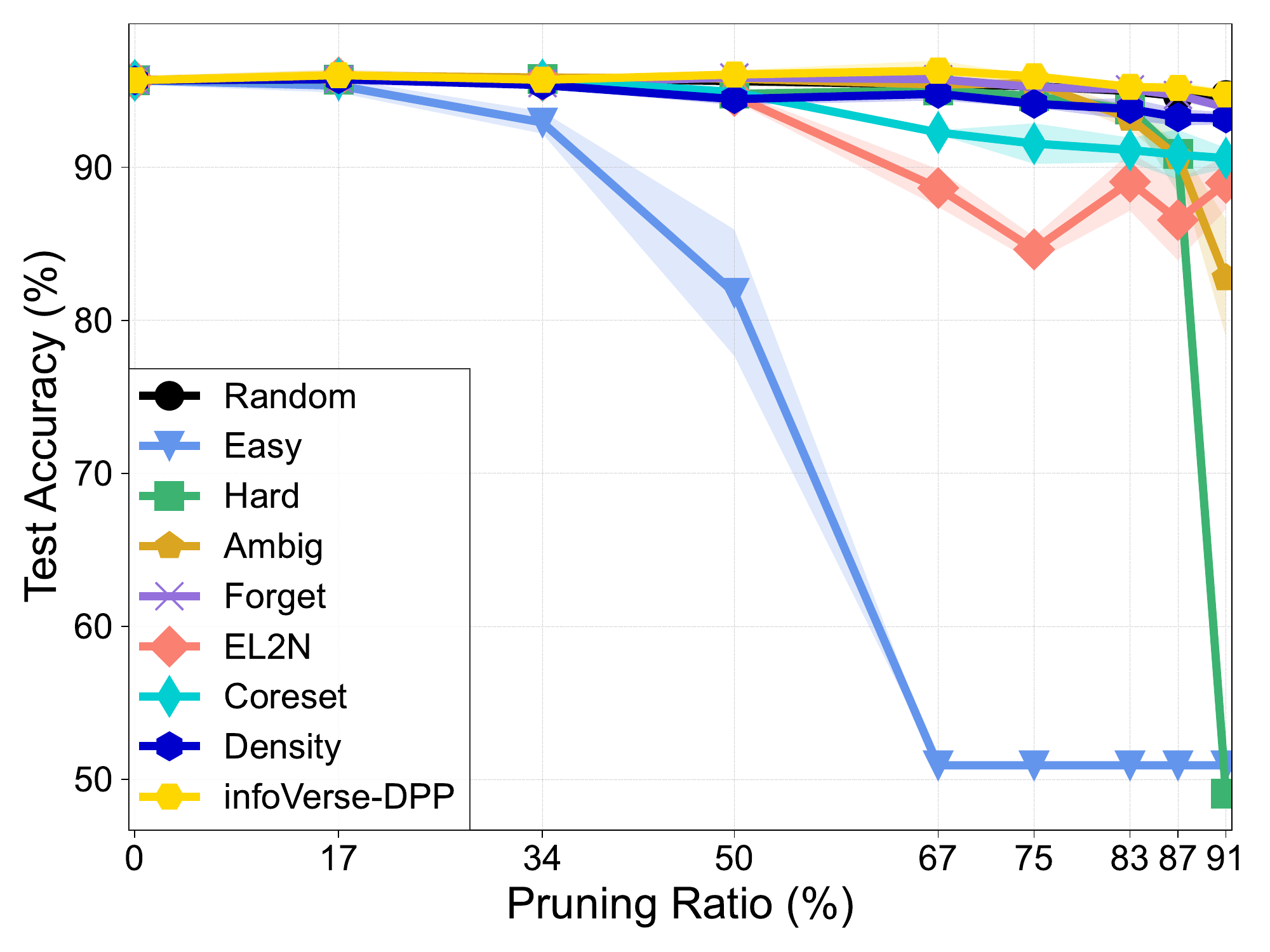}
        \label{fig:app_ds3}
    }
    \vspace{-2mm}
    \subfigure[RTE]
    {
        \includegraphics[width=0.43\textwidth]{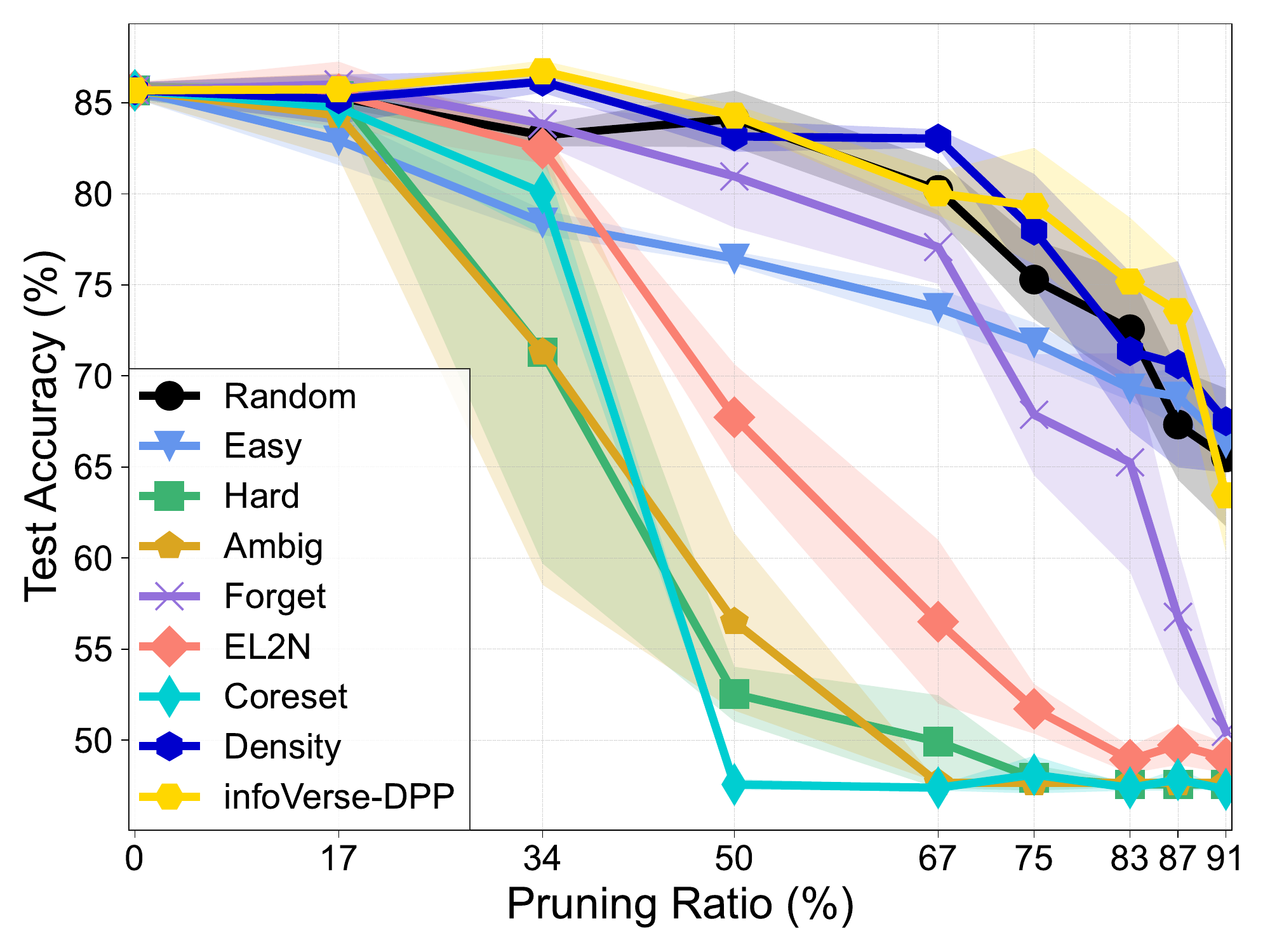}
        \label{fig:app_ds4}
    }
    \vspace{-2mm}
    \subfigure[QNLI]
    {
        \includegraphics[width=0.43\textwidth]{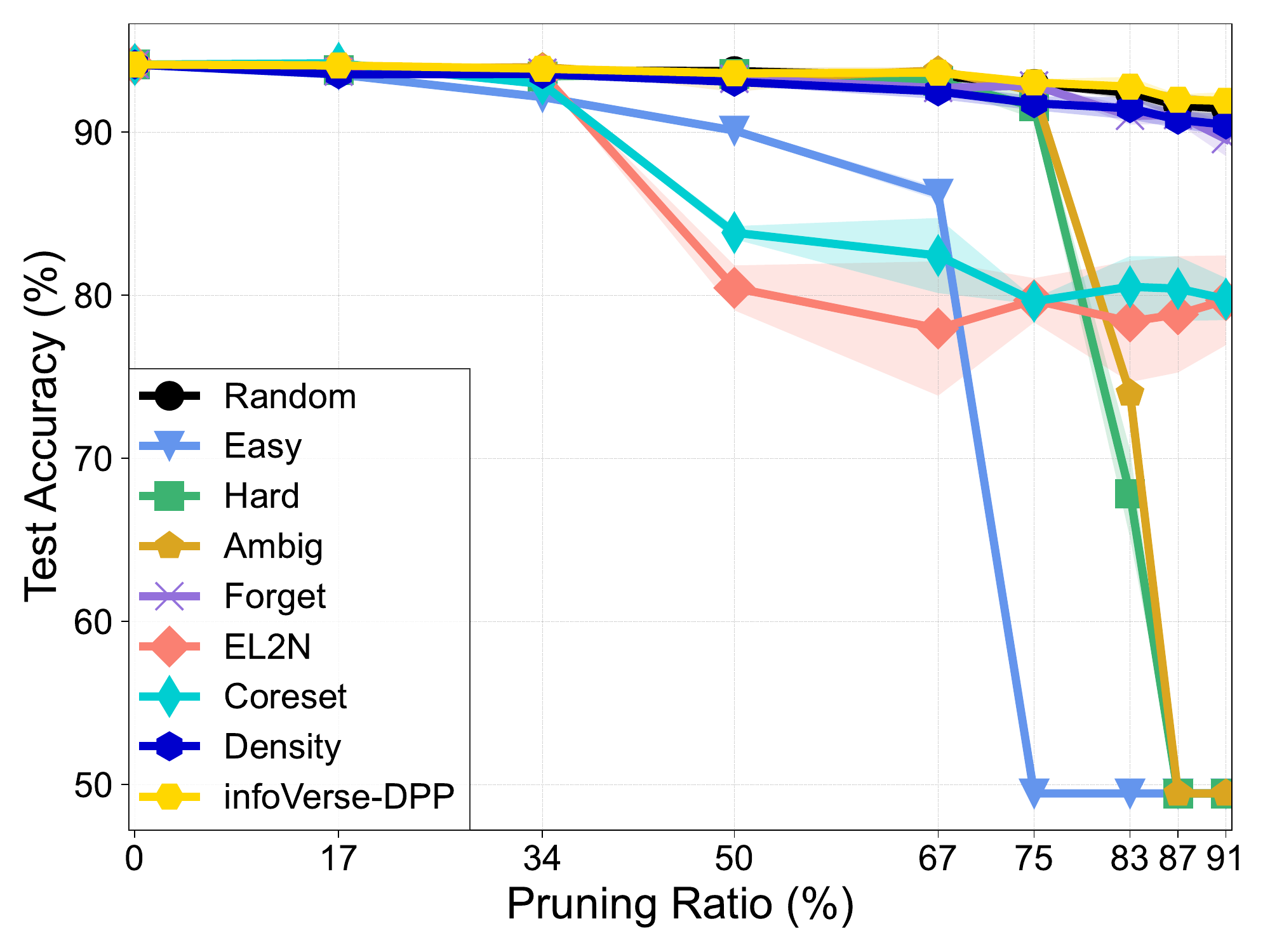}
        \label{fig:app_ds5}
    }
}
\vspace{-2mm}
\end{center}
\caption{Data pruning performance on different datasets: (a) CoLA, (b) SST-2, (C) RTE, and (d) QNLI.}
\label{fig:app_fig4}
\end{figure}

%% file: tables/acl23_table8_info_dpp_ablation.tex
\begin{table}[t]
	\begin{center}
	\caption{Ablation study with each component of \name{}-DPP. Average test accuracy of finetuned RoBERTa-large classifiers over 8 different pruning ratio on WinoGrande and CoLA are compared, similar to Table \ref{table:pruning}.}
	\label{table:ablation_info_dpp}
    \begin{adjustbox}{width=0.95\linewidth}
	\begin{tabular}{cc|cc}
		\toprule
		Feature & Sampling & WinoGrande & CoLA
		              \\ \midrule
		Random & Random & {73.1}\ms{0.09} & {60.7}\ms{0.63} \\
		Classifier & Coreset & {72.2}\ms{0.91} & {55.5}\ms{0.93} \\
		Classifier & DPP & {73.5}\ms{0.25} & {62.1}\ms{0.19} \\
		\name{} & Coreset & 71.9\ms{0.47} & {61.0}\ms{0.26}\\ 
		\name{} & DPP & \textbf{{74.6}}\ms{0.24} & \textbf{{62.5}}\ms{0.14} \\\bottomrule
	\end{tabular}
    \end{adjustbox}
    \end{center}
\end{table}

%% file: figures/acl23_figure14.tex
\begin{figure}[t]
\begin{center}
{
    \subfigure[CoLA]
    {
        \includegraphics[width=1.0\columnwidth, trim={1.0cm 0.5cm 0.5cm 2cm},clip]{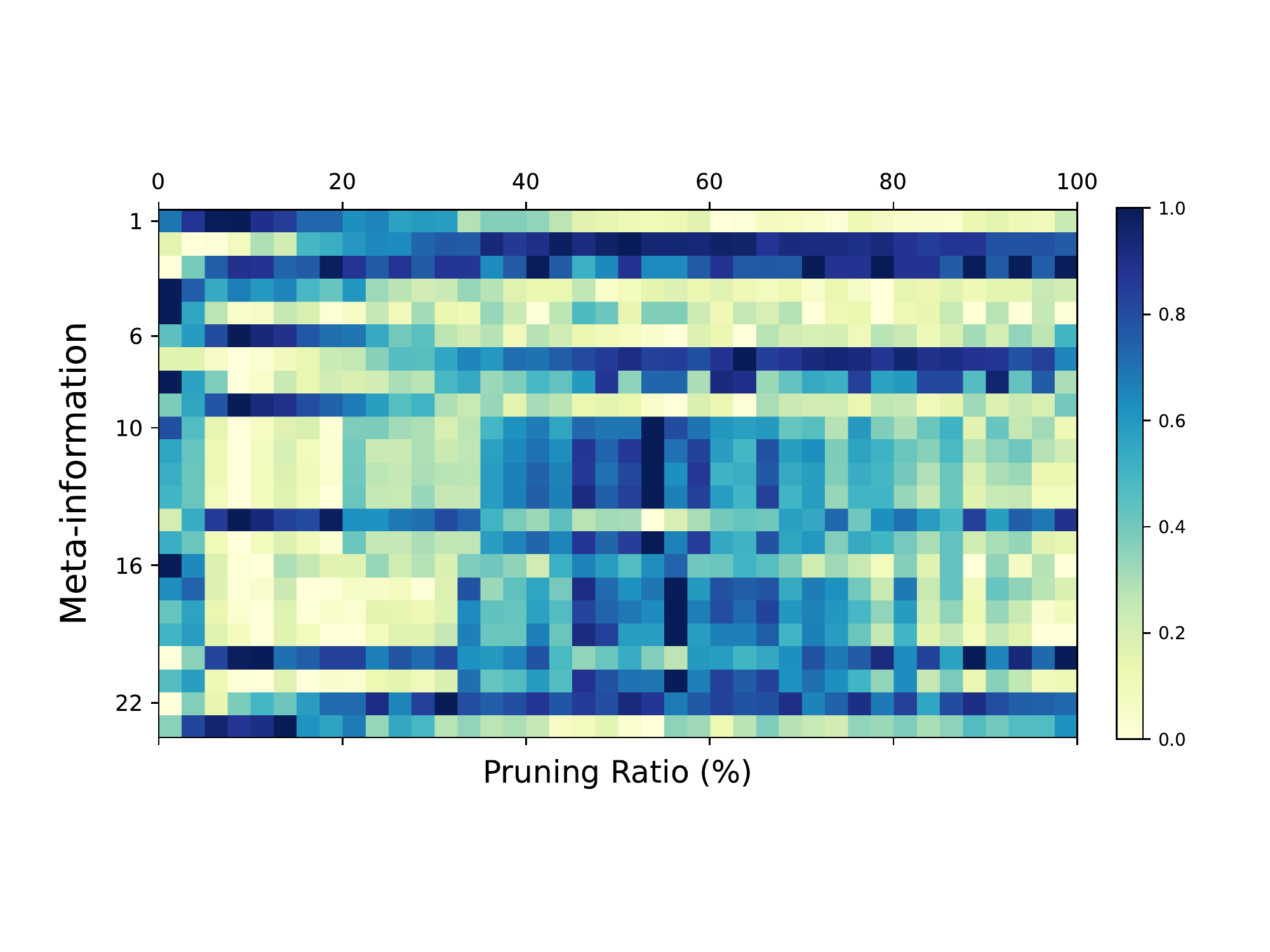}
        \label{fig:app_dp_analysis_cola}
    }
    \vspace{-5mm}
    \subfigure[WinoGrande]
    {
        \includegraphics[width=1.0\columnwidth, trim={1.0cm 0.5cm 0.5cm 2cm},clip]{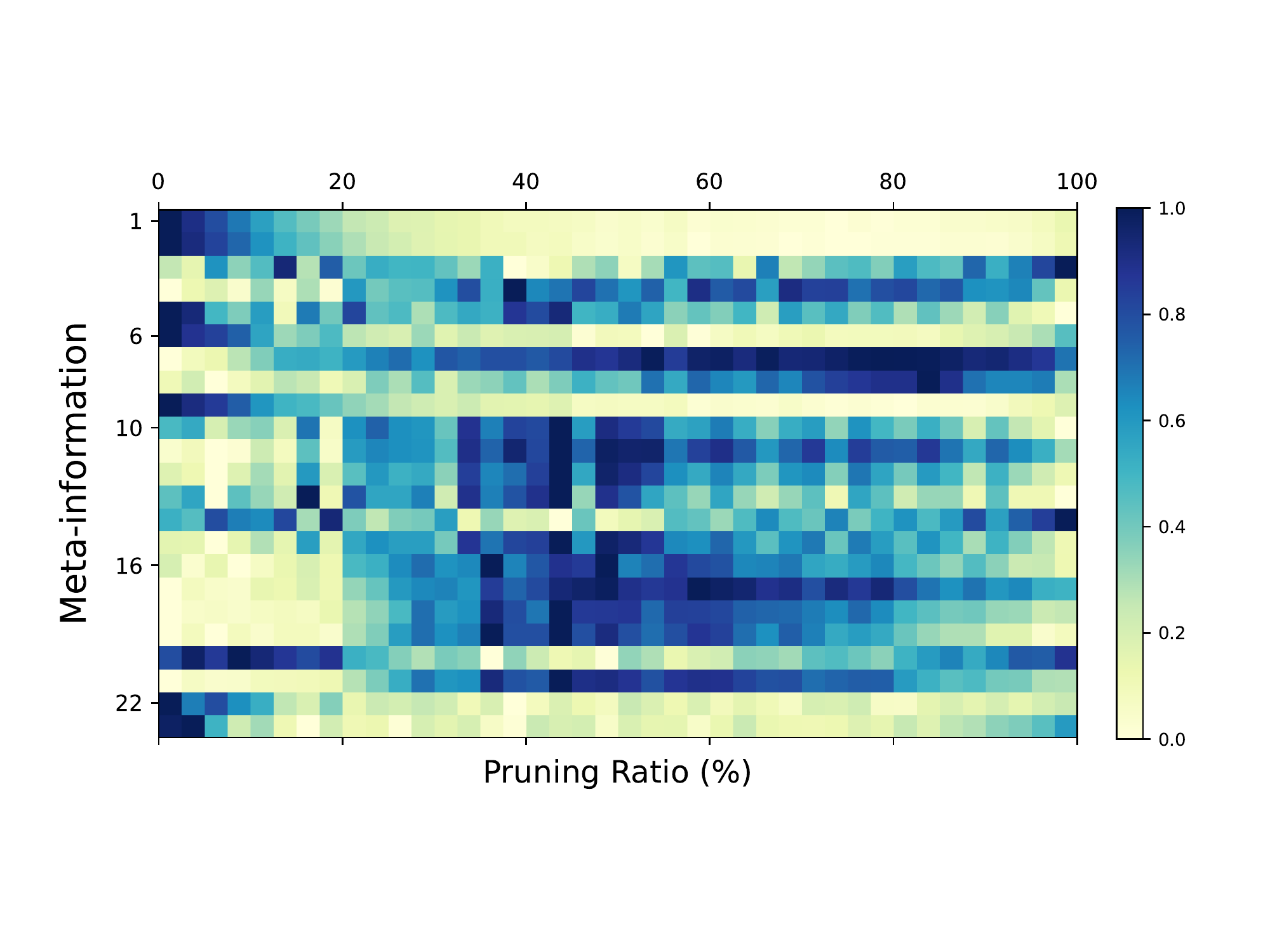}
        \label{fig:app_dp_analysis_wino}
    }
    \vspace{-5mm}
    \subfigure[RTE]
    {
        \includegraphics[width=1.0\columnwidth, trim={1.0cm 0.5cm 0.5cm 2cm},clip]{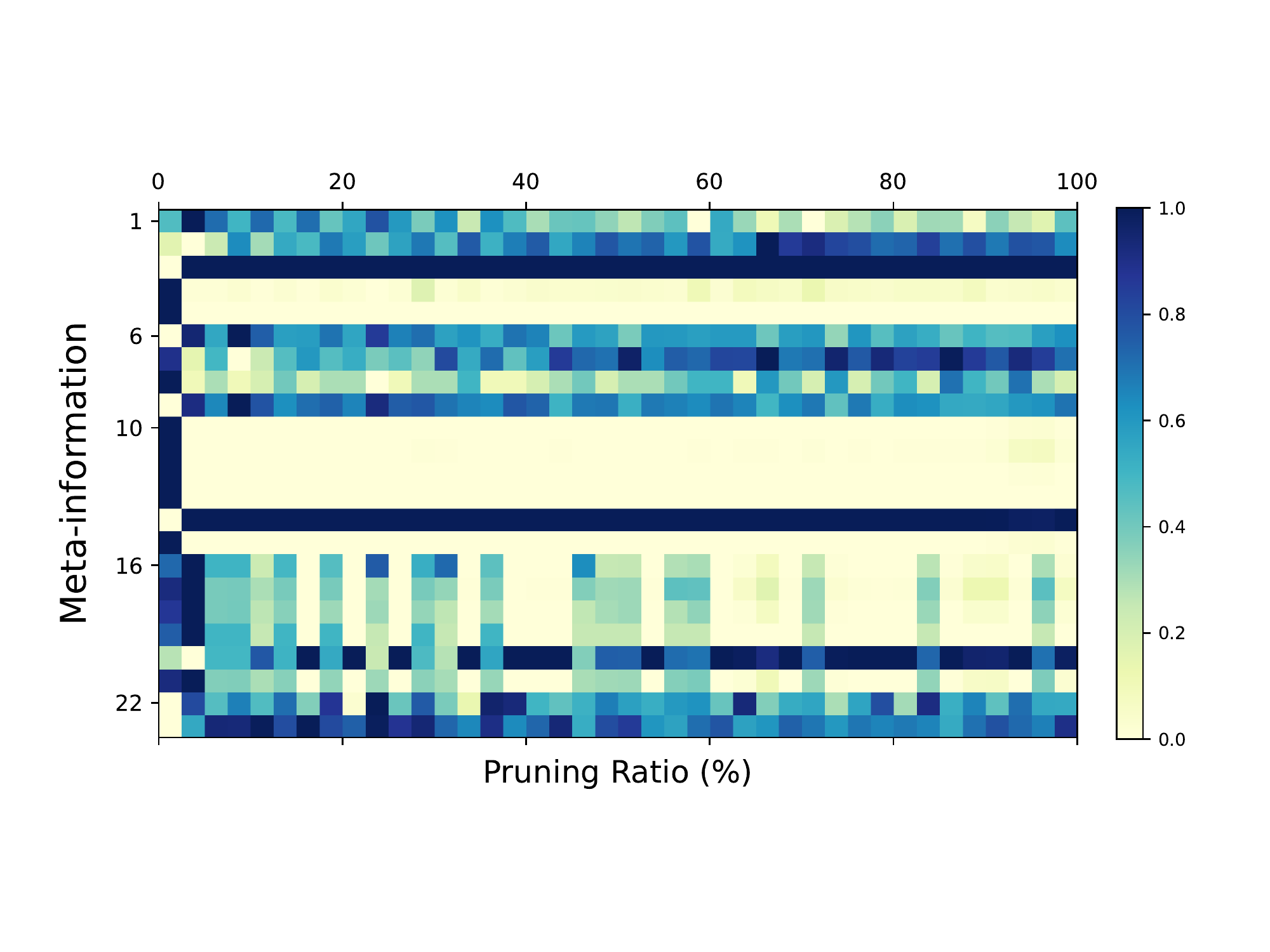}
        \label{fig:app_dp_analysis_rte}
    }
    \vspace{-5mm}
    \subfigure[QNLI]
    {
        \includegraphics[width=1.0\columnwidth, trim={1.0cm 0.5cm 0.5cm 2cm},clip]{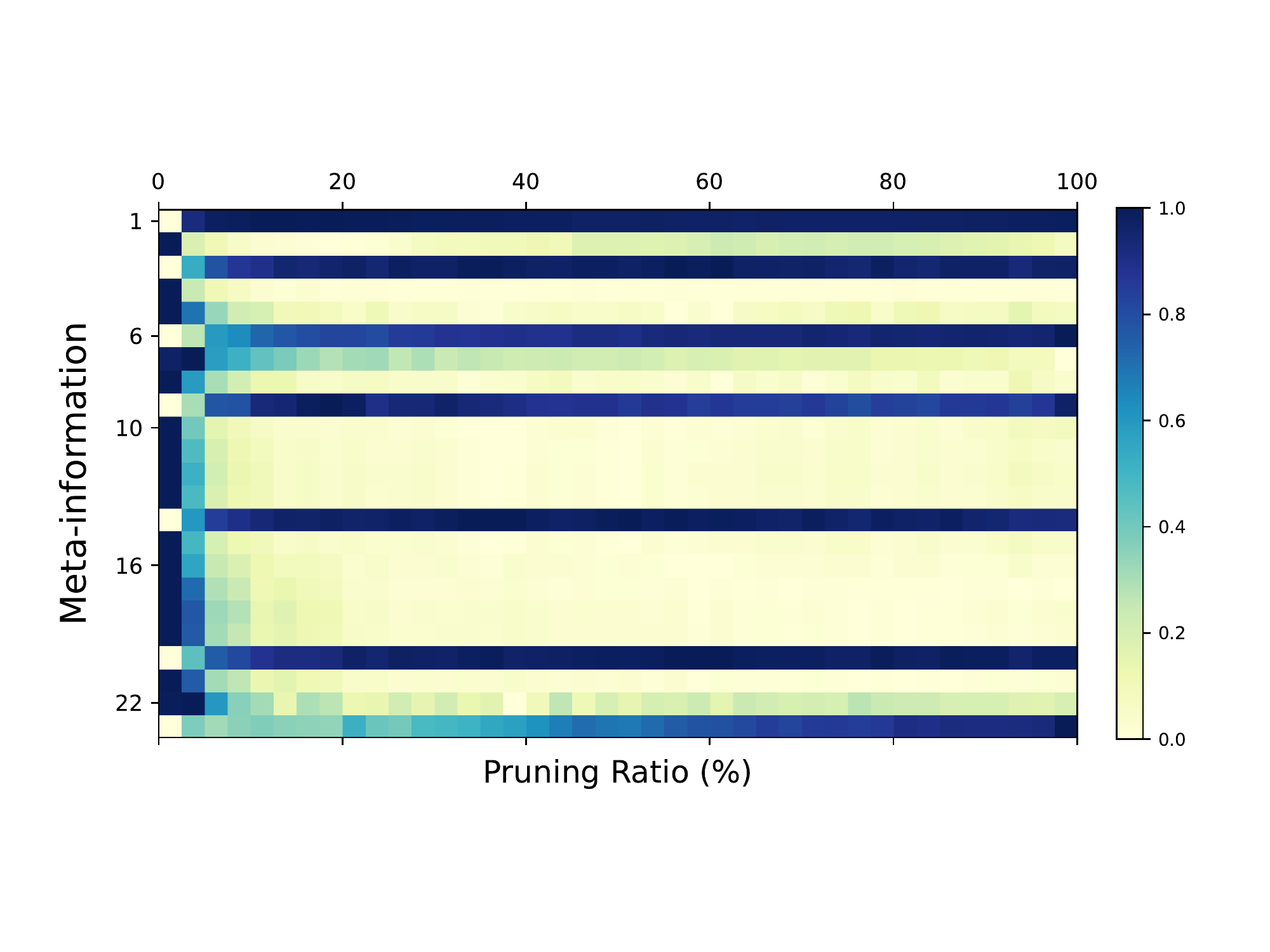}
        \label{fig:app_dp_analysis_qnli}
    }
}
\vspace{-2mm}
\end{center}
\caption{Dynamics of data pruning with \name{}-DPP on different datasets: (a) CoLA, (b) WinoGrande, (C) RTE, and (d) QNLI.}
\label{fig:app_pruning_analysis}
\end{figure}

%% file: figures/acl23_figure15.tex
\begin{figure}[t]
\begin{center}
{
    \subfigure[\name{}-DPP]
    {
        \includegraphics[width=1.0\columnwidth, trim={1.0cm 0.5cm 0.5cm 2cm},clip]{figures/final_figures/acl23_pruning_ablation_sst2.pdf}
    }
    \vspace{-2mm}
    \subfigure[Variability]
    {
        \includegraphics[width=1.0\columnwidth, trim={1.0cm 0.5cm 0.5cm 2cm},clip]{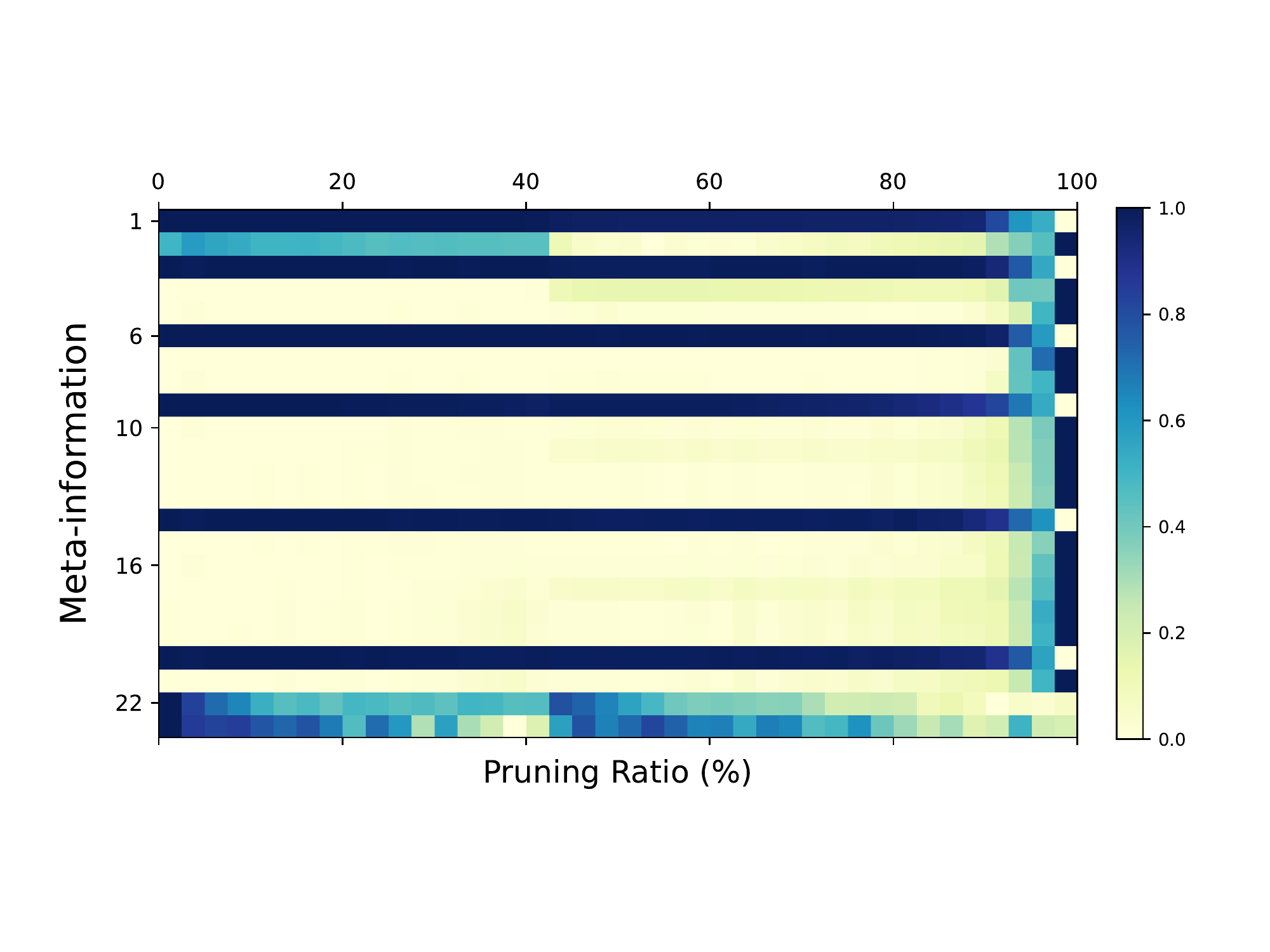}
        \label{fig:app_dp_analysis_variab}
    }
    \vspace{-2mm}
    \subfigure[Random]
    {
        \includegraphics[width=1.0\columnwidth, trim={1.0cm 0.5cm 0.5cm 2cm},clip]{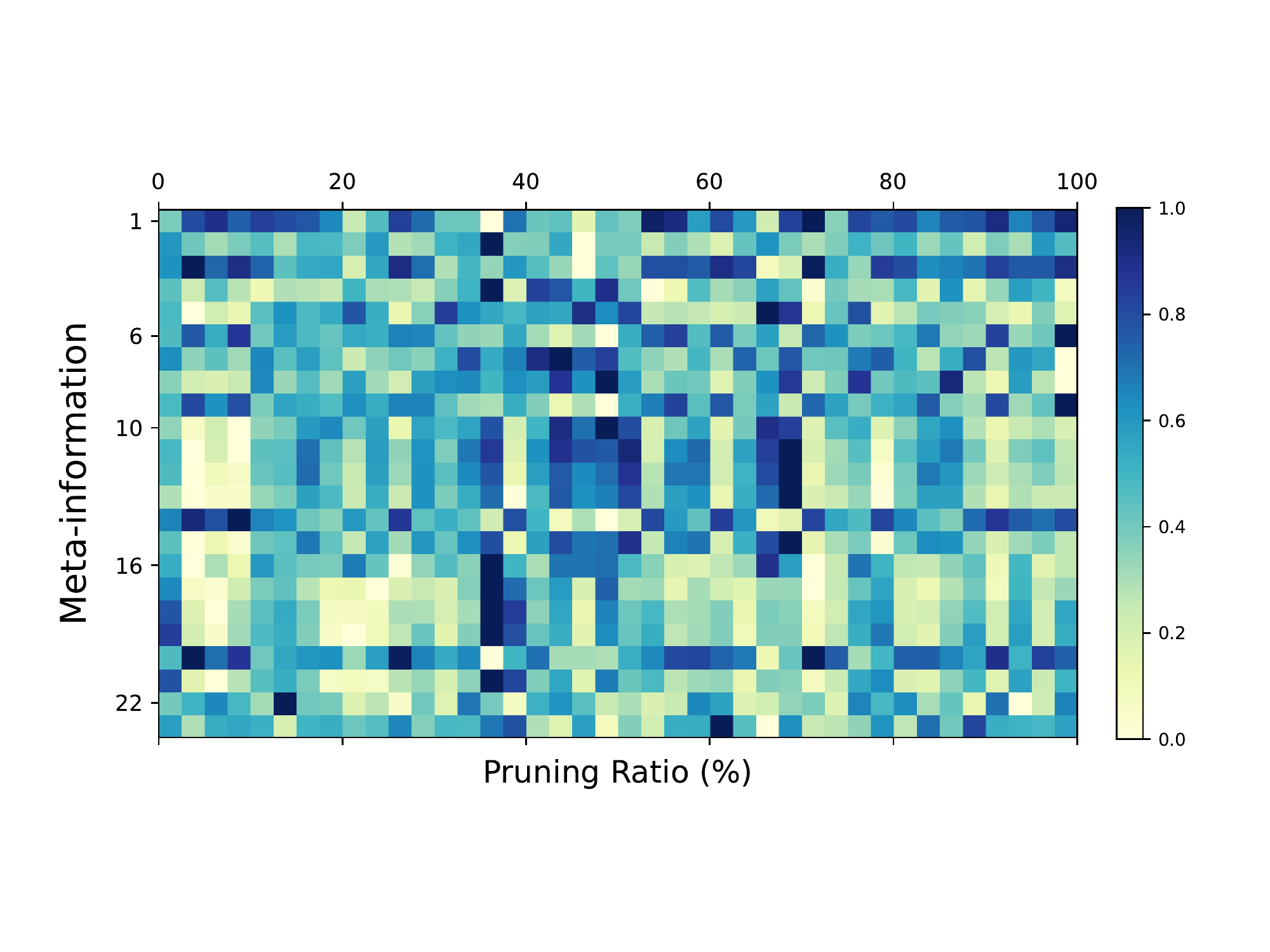}
        \label{fig:app_dp_analysis_rand}
    }
}
\vspace{-2mm}
\end{center}
\caption{Dynamics of data pruning with different selection method on SST-2 dataset: (a) \name{}-DPP, (b) Variability, and (c) Random.}
\label{fig:app_pruning_analysis2}
\end{figure}

%% file: figures/acl23_figure16.tex
\begin{figure}[t]
\begin{center}
{   
    \vspace{-3mm}
    \subfigure[WinoGrande]
    {
        \includegraphics[width=0.36\textwidth, trim={4.0cm 1.5cm 1.5cm 0cm},clip]{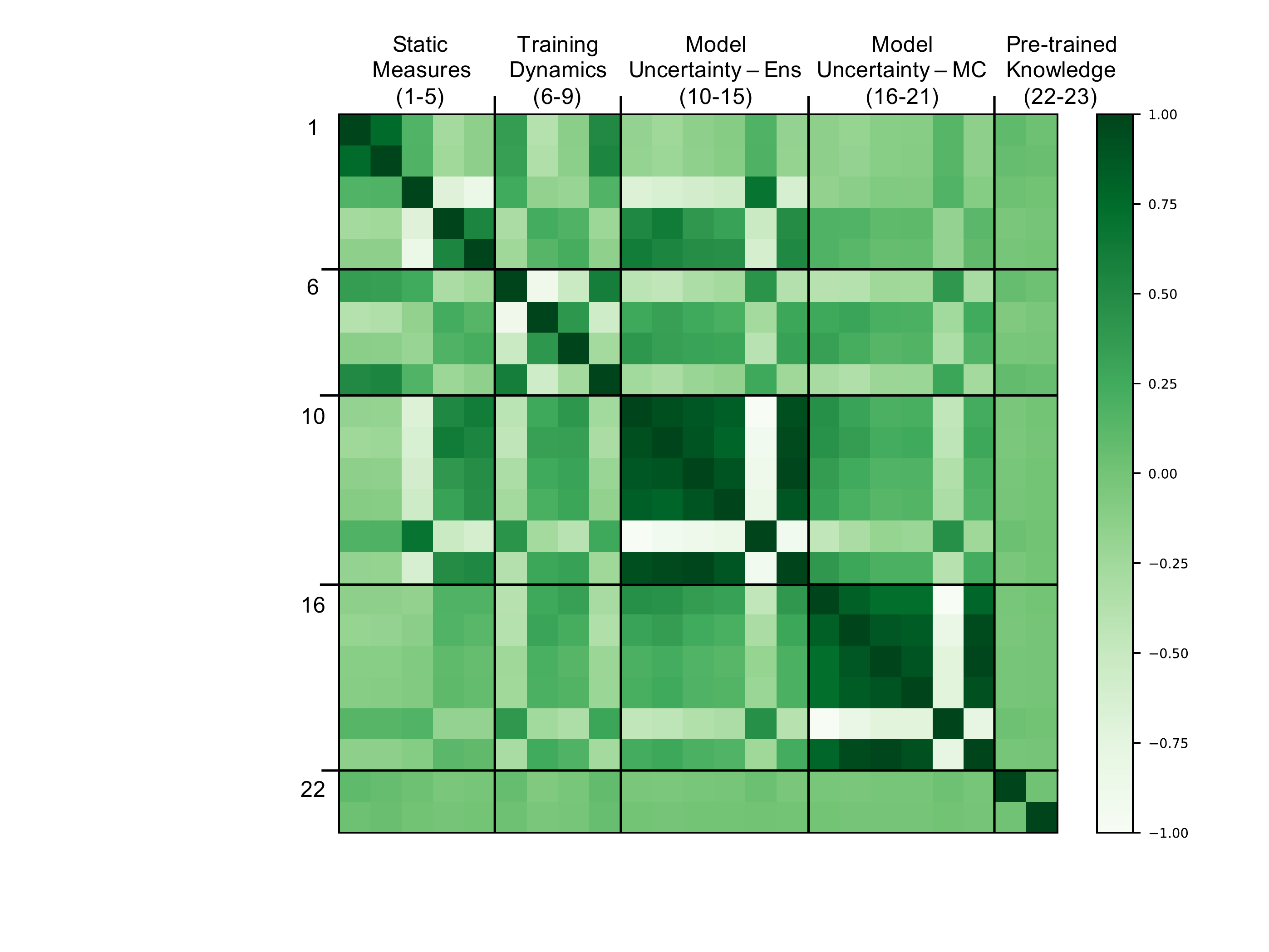}
        \label{fig:app_conf_wino}
    }
    \vspace{-3mm}
    \subfigure[SST-2]
    {
        \includegraphics[width=0.36\textwidth, trim={4.0cm 1.5cm 1.5cm 0cm},clip]{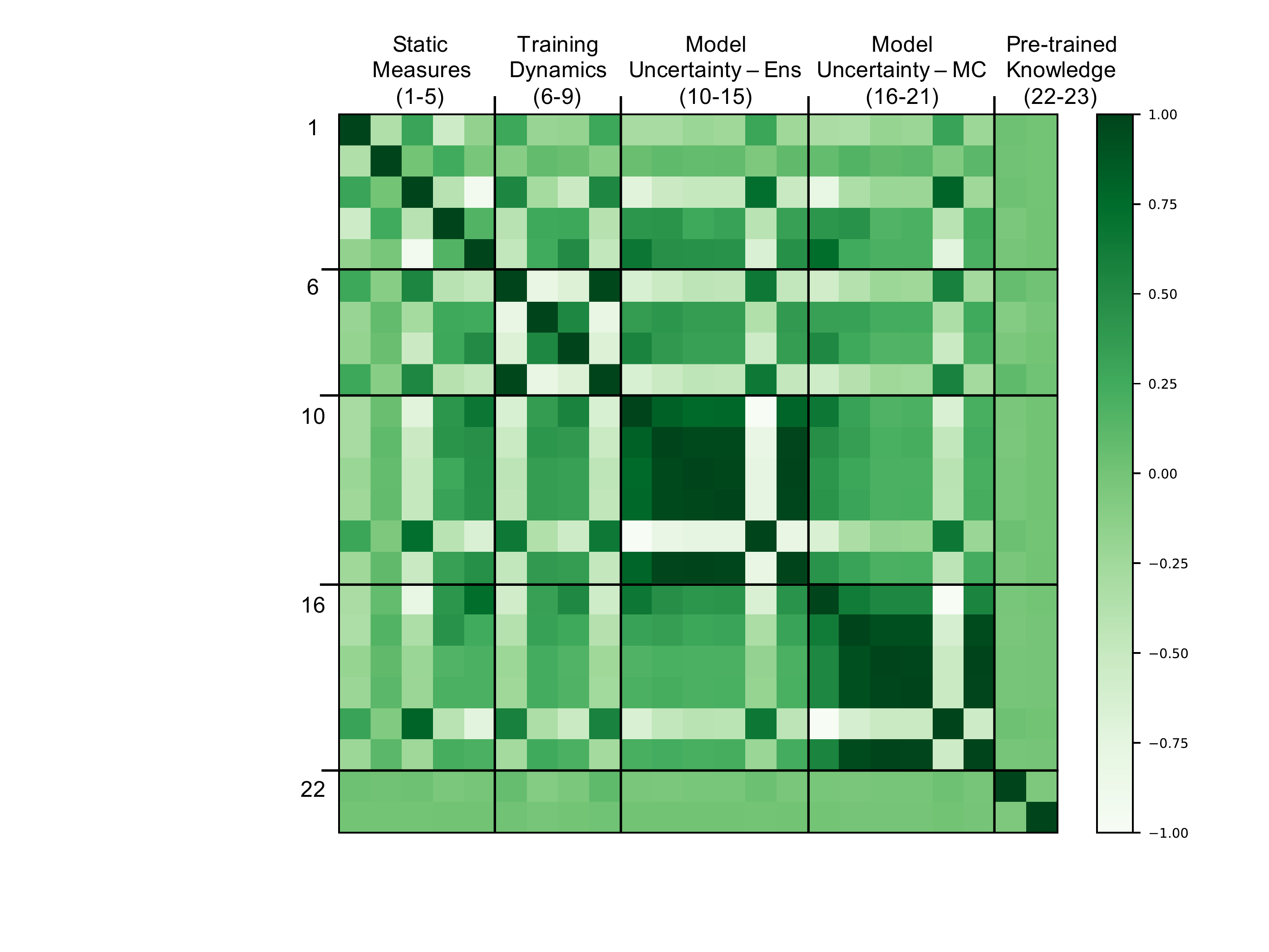}
        \label{fig:app_conf_sst2}
    }
    \vspace{-3mm}
    \subfigure[CoLA]
    {
        \includegraphics[width=0.36\textwidth, trim={4.0cm 1.5cm 1.5cm 0cm},clip]{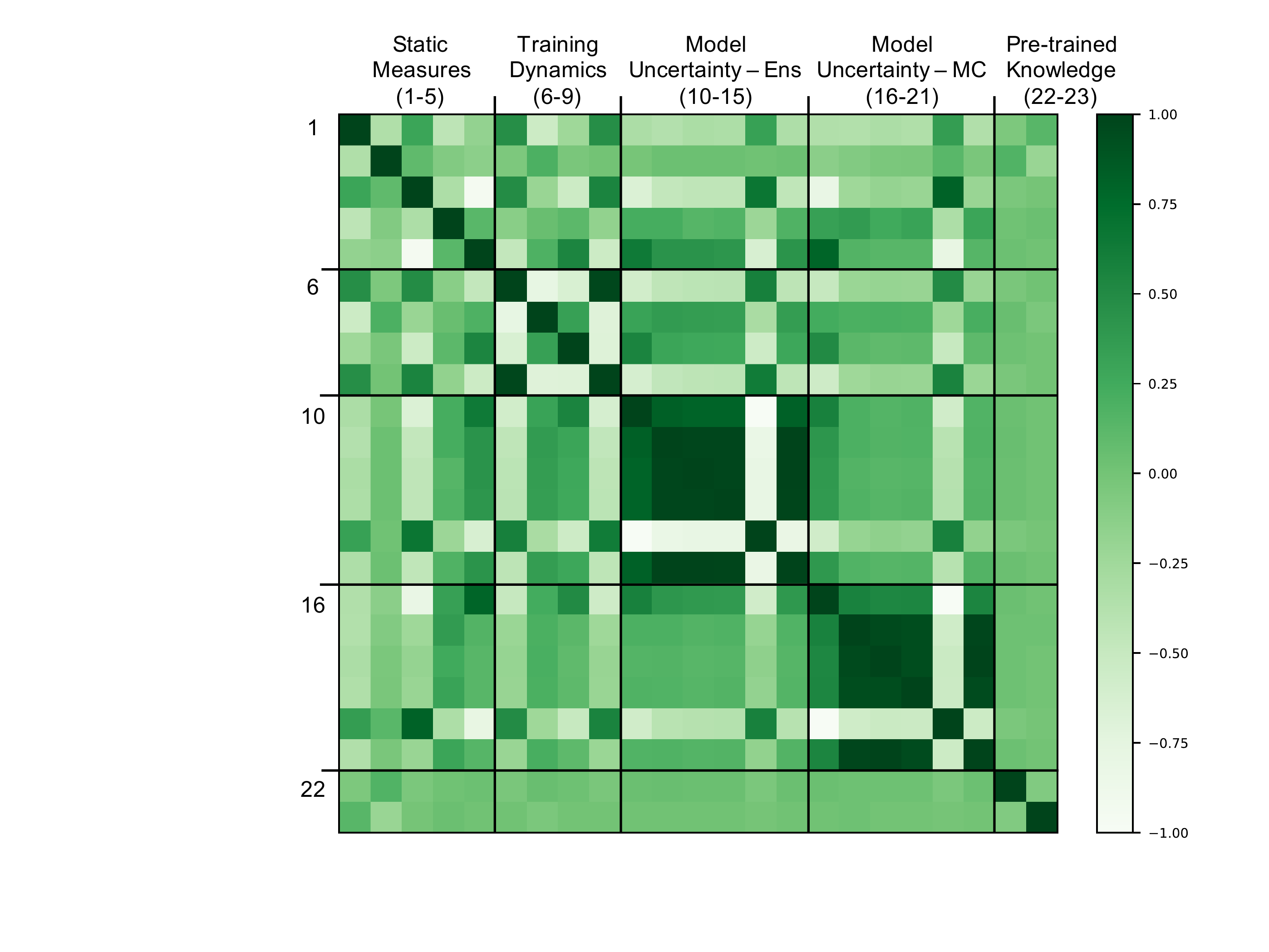}
        \label{fig:app_conf_cola}
    }
    \vspace{-3mm}
    \subfigure[RTE]
    {
        \includegraphics[width=0.36\textwidth, trim={4.0cm 1.5cm 1.5cm 0cm},clip]{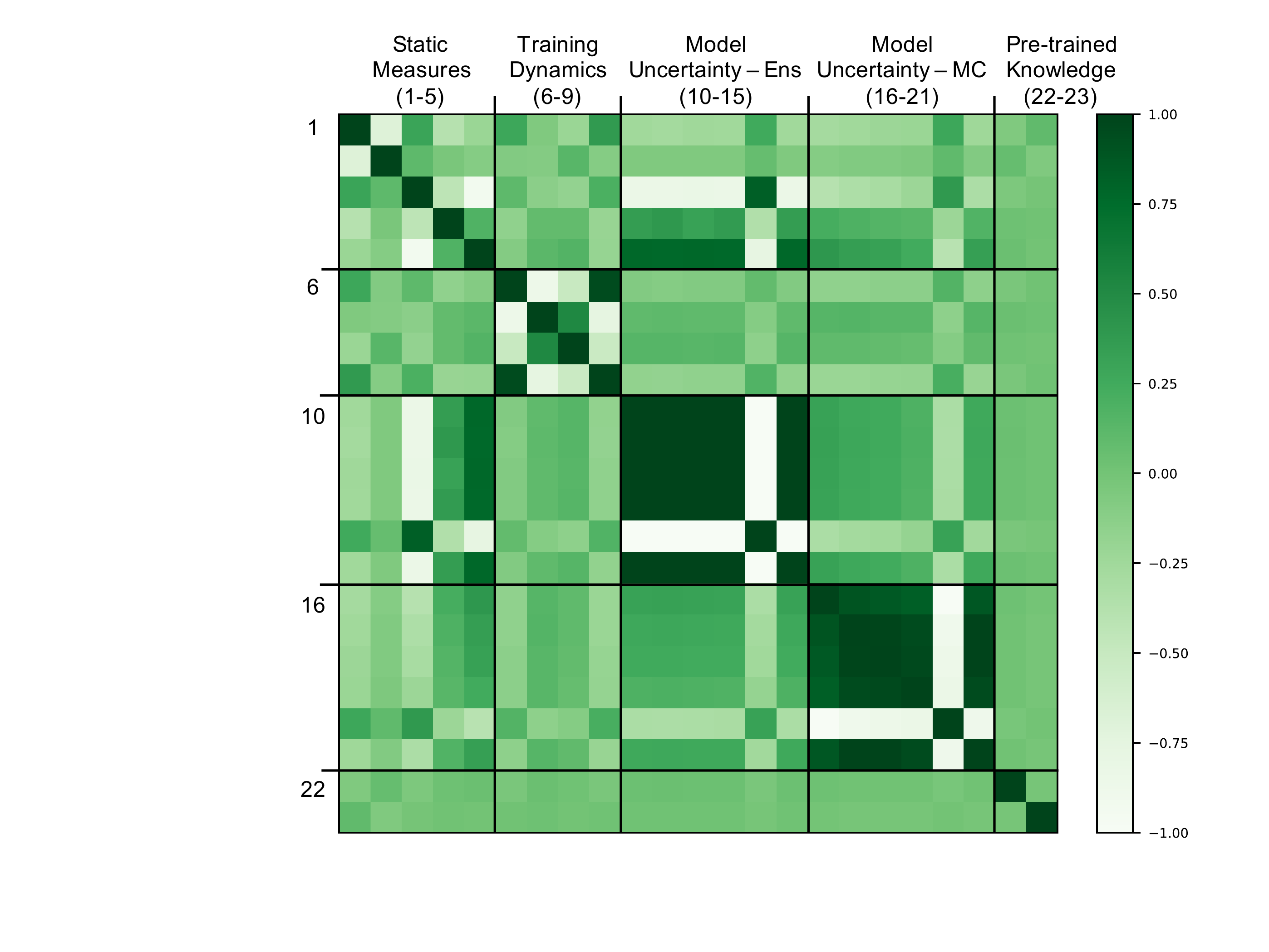}
        \label{fig:app_conf_rte}
    }
}
\vspace{-3mm}
\end{center}
\caption{Confusion matrices between meta-information on different datasets: (a) WinoGrande, (b) SST-2, (C) CoLA, and (d) RTE.}
\label{fig:app_conf_pruning}
\end{figure}

%% file: figures/acl23_figure17.tex
\begin{figure*}[t]
	\centering
	\includegraphics[width=1.0\linewidth]{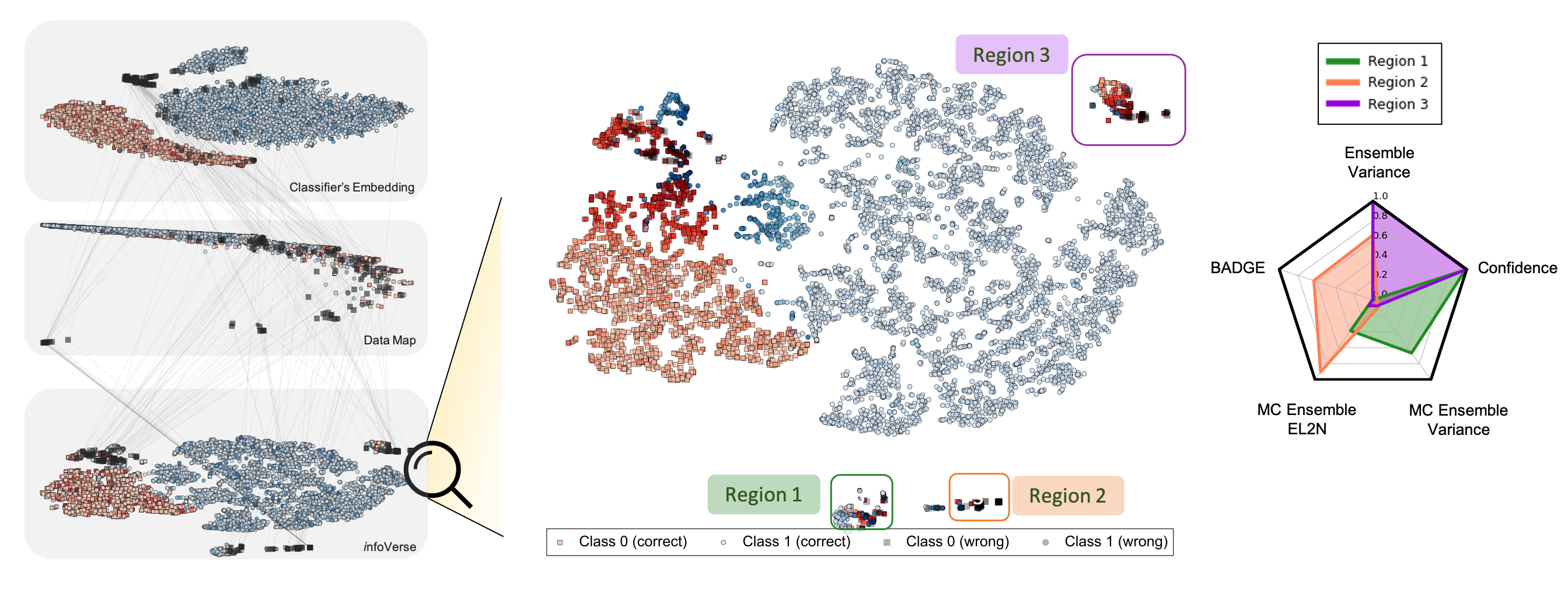}\vspace{-3mm}
    \caption{
    \name{} (bottom left) on CoLA along with other feature spaces: classifier embedding (top left) and \textit{data map} \cite{swayamdipta2020cartography} (middle left). (middle) Zoomed version of \name{} is presented. (right) Score distribution of each wrong region characterized by \name{}.
    }
    \label{fig:figure2_cola}
\end{figure*}

%% file: figures/acl23_figure18.tex
\begin{figure*}[t]
	\centering
	\includegraphics[width=1.0\linewidth]{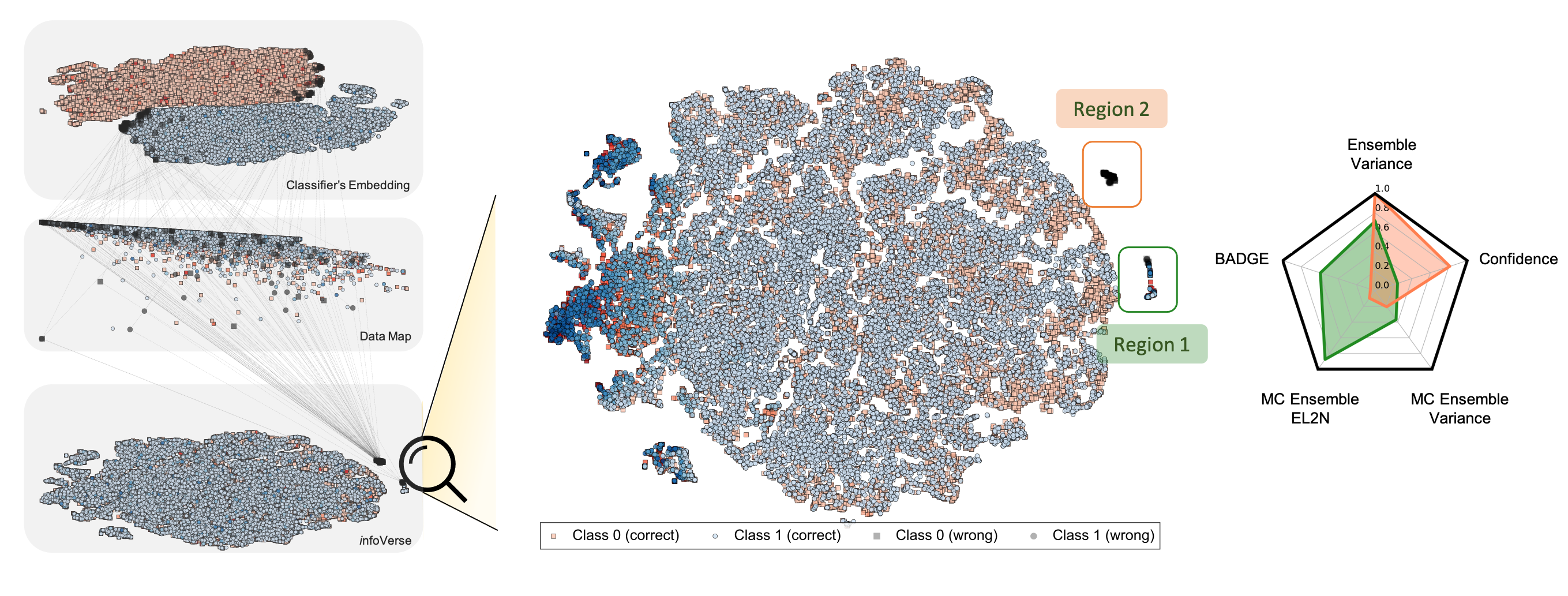}\vspace{-3mm}
    \caption{
    \name{} (bottom left) on WinoGrande along with other feature spaces: classifier embedding (top left) and \textit{data map} \cite{swayamdipta2020cartography} (middle) Zoomed version of \name{} is presented. (right) Score distribution of each wrong region characterized by \name{}.
    }
    \label{fig:figure2_wino}
\end{figure*}

%% file: figures/acl23_figure19.tex
\begin{figure*}[t]
	\centering
	\includegraphics[width=1.0\linewidth]{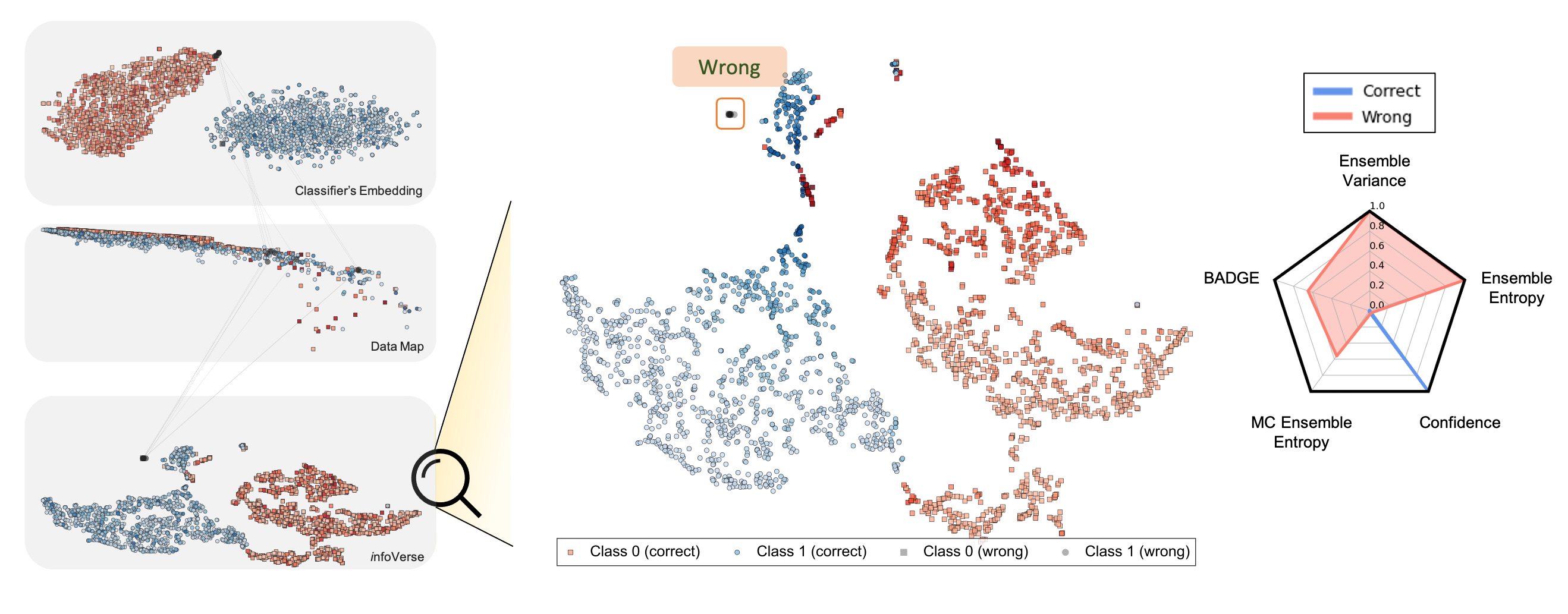}\vspace{-3mm}
    \caption{
    \name{} (bottom left) on RTE along with other feature spaces: classifier embedding (top left) and \textit{data map} \cite{swayamdipta2020cartography} (middle left). (middle) Zoomed version of \name{} is presented. (right) Score distribution of each wrong region characterized by \name{}.
    }
    \label{fig:figure2_rte}
\end{figure*}

%% file: figures/acl23_figure20.tex
\begin{figure*}[t]
	\centering
	\includegraphics[width=1.0\linewidth]{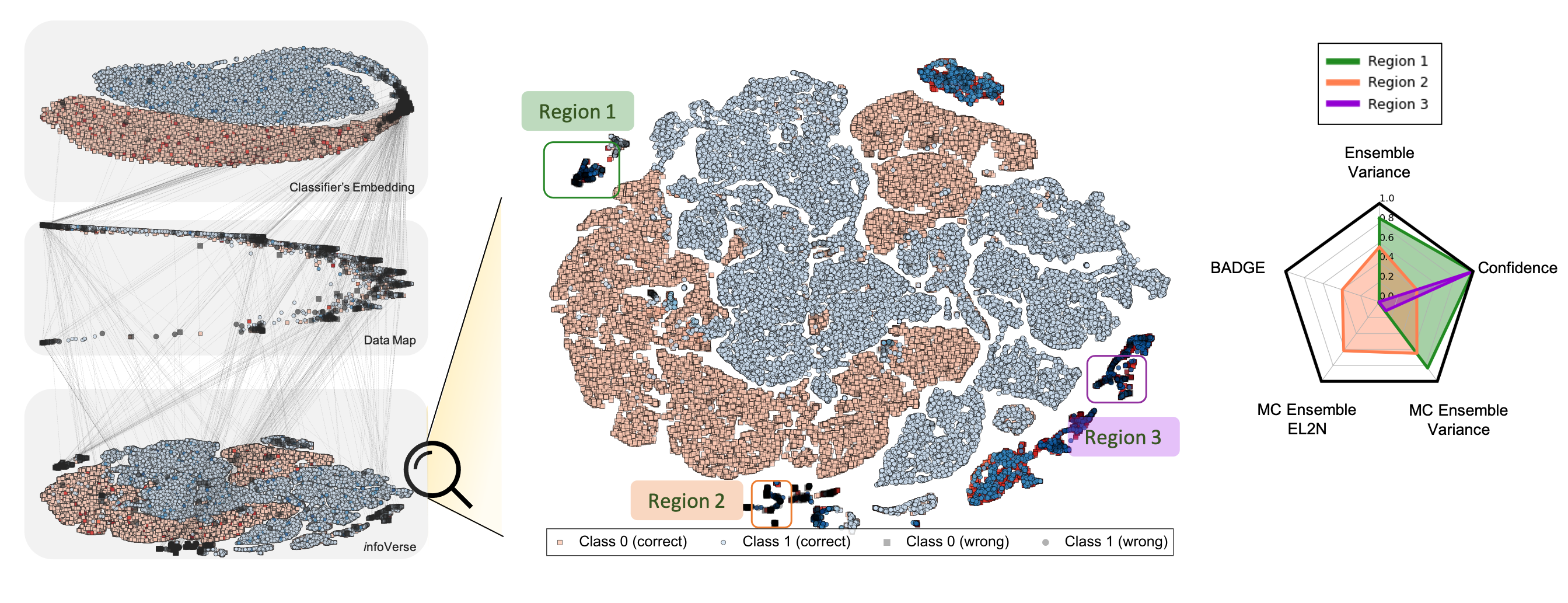}\vspace{-3mm}
    \caption{
    \name{} (bottom left) on SST-2 along with other feature spaces: classifier embedding (top left) and \textit{data map} \cite{swayamdipta2020cartography} (middle left). (middle) Zoomed version of \name{} is presented. (right) Score distribution of each wrong region characterized by \name{}.
    }
    \label{fig:figure2_sst}
\end{figure*}

%% file: tables/acl23_table9_ablation_multiple.tex
\begin{table}[t]
    \caption{Detection accuracy (\%) of linear classifier trained with multiple meta-information.\vspace{-3mm}}
	\begin{center}
	\begin{adjustbox}{width=1.0\columnwidth}
	\begin{tabular}{r|cccc}
 		\toprule
		Feature Space & Mis-pred & Mis-labeled & OOD & Adv \\ \midrule
		Classifier Embedding    & {10.4} & {89.1} & {89.0} & {85.9} \\ \midrule
		$^{*}$\name{}     & {97.5} & {94.2} & {90.3} & {87.1} \\ 
		\name{}     & \textbf{99.9} & \textbf{94.3} & \textbf{91.4} & \textbf{87.5} \\ 
		- (4)       & {99.8} & {94.2} & {86.3} & {86.0} \\ 
		- (3)-\textit{MC}       & {85.7} & {94.2} & {82.9} & {86.0} \\ 
		- (3)-\textit{Ens}       & {77.9} & {94.2} & {78.6} & {86.0} \\ 
		- (2)       & {49.9} & {94.1} & {69.9} & {86.0} \\ 
 		\bottomrule
	\end{tabular}
    \end{adjustbox}
    \end{center}
    \label{table:ablation}
    \vspace{-4mm}
\end{table}